\documentclass[conf]{new-aiaa}
\usepackage[utf8]{inputenc}
\usepackage{textcomp}

\usepackage[]{graphicx}
\graphicspath{
    {./figures/}
    {./figures/01_intro/}
    {./figures/02_related_work/}
    {./figures/04_spnv3/}
    {./figures/05_ood/}
}
\DeclareGraphicsExtensions{.pdf,.eps,.jpeg,.png}

\usepackage{subcaption}
\usepackage{amsmath, amsfonts}
\usepackage{longtable, tabularx, booktabs}
\usepackage{makecell, multirow}
\usepackage{tikz}
\usepackage{bm}

\usepackage{cleveref}

\crefname{equation}{Eq.}{Eqs.}
\Crefname{equation}{Equation}{Equations}
\crefname{figure}{Fig.}{Figs.}
\Crefname{figure}{Figure}{Figures}
\crefname{table}{Table}{Tables}
\Crefname{table}{Table}{Tables}
\crefname{algorithm}{Algorithm}{Algorithms}
\Crefname{algorithm}{Algorithm}{Algorithms}
\crefname{appendix}{Appendix}{Appendices}
\Crefname{appendix}{Appendix}{Appendices}

\crefformat{section}{Section #2#1#3}
\crefmultiformat{section}{Sections #2#1#3}{, #2#1#3}{, #2#1#3}{, #2#1#3}
\crefrangeformat{section}{Sections #3#1#4--#5#2#6}

\newcommand{\tikzcircle}[1]{\tikz[baseline=-0.5ex]\draw[#1,fill=#1,radius=1.5pt] (0,0) circle ;}%

\newcommand{\synthetic}{\texttt{synthetic}}
\newcommand{\lightbox}{\texttt{lightbox}}
\newcommand{\sunlamp}{\texttt{sunlamp}}
\newcommand{\prisma}{\texttt{prisma25}}

\newcommand{\norm}{\textsf{NORM}}

\newcommand{\mnstd}[2]{#1\tiny{(#2)}}

\usepackage{changes}
\makeatletter
\AddToHook{cmd/added/before}{\def\Changes@AuthorColor{red}}
\AddToHook{cmd/deleted/before}{\def\Changes@AuthorColor{blue}}
\makeatother

\title{Bridging the Domain Gap for Flight-Ready Spaceborne Vision}

\author{Tae~Ha~Park\footnote{GN\&C Engineer, 632 Gukhoe-daero, Yeongdeungpo-gu; thpark@naraspace.com. Member AIAA.}}
\affil{Nara Space Technology Inc., Seoul, 07245, Republic of Korea}

\author{Simone~D'Amico\footnote{Associate Professor, Department of Aeronautics \& Astronautics, 496 Lomita Mall; damicos@stanford.edu. Associate Fellow AIAA.}}
\affil{Stanford University, Stanford, CA, 94305, USA}

\begin{document}

\maketitle

\footnotetext{This work was done as part of the Ph.D.~program of T.~H.~Park at the Space Rendezvous Laboratory (SLAB), Stanford University.}

\begin{abstract}
This work presents Spacecraft Pose Network v3 (SPNv3), a Neural Network (NN) for monocular pose estimation of a known, non-cooperative target spacecraft. SPNv3 is designed and trained to be computationally efficient while providing robustness to spaceborne images that have not been observed during offline training and validation on the ground. These characteristics are essential to deploying NNs on space-grade edge devices. They are achieved through careful NN design choices, and an extensive trade-off analysis reveals features such as data augmentation, transfer learning and vision transformer architecture as a few of those that contribute to simultaneously maximizing robustness and minimizing computational overhead. Experiments demonstrate that the final SPNv3 can achieve state-of-the-art pose accuracy on hardware-in-the-loop images from a robotic testbed while having trained exclusively on computer-generated synthetic images, effectively bridging the domain gap between synthetic and real imagery. At the same time, SPNv3 runs well above the update frequency of modern satellite navigation filters when tested on a representative graphical processing unit system with flight heritage. Overall, SPNv3 is an efficient, flight-ready NN model readily applicable to close-range rendezvous and proximity operations with target resident space objects.
\end{abstract}




\section{Introduction}
\label{sec:01_intro}

\lettrine{A}{utonomous} Rendezvous, Proximity Operations and Docking (RPOD) capability with non-cooperative Resident Space Objects (RSO) is a core technological requirement for various future space missions. Aimed at sustainable space development, these missions include on-orbit servicing and refueling operations such as the now-canceled OSAM missions by NASA \citep{nasa_osam_1, nasa_osam_2} and active debris removal such as RemoveDEBRIS \citep{aglietti_2019_aer_removedebris} by Surrey Space Center, ADRAS-J \citep{adras_j} by Astroscale and ClearSpace-1 \citep{clearspace-1} by ClearSpace SA. The RPOD capability with such targets---defunct satellites, debris, etc.---requires accurate, real-time knowledge of the position and orientation (i.e., \emph{pose}) of the target spacecraft with respect to the servicer spacecraft. In these scenarios, the \emph{non-cooperative} nature of the target implies that it is without active communication links or aiding fiduciary markers on its surface to guide the rendezvous process, such that the servicer must be able to estimate and track the target’s pose using the onboard sensors and computers only. Given such constraints, a monocular camera is an attractive choice of sensor due to its low Size-Weight-and-Power-Cost (SWaP-C), which is suitable for miniaturized systems such as SmallSats and CubeSats. Moreover, low SWaP-C implies additional sensor redundancy and fail-proofing compared to more complex sensor systems such as LIght Detection And Ranging (LIDAR) and stereovision. In particular, LIDAR allows accurate range detection at far range \citep{ruel_2012_jfr_tridar, christian_2013_gnc_lidarreview}, but it is subject to much stricter SWaP-C requirements. On the other hand, while stereovision utilizes low SWaP-C vision sensors amenable to SmallSats \citep{oumer_2012_icpr_stereo, fourie_2014_jsr_sphere}, its applicability for range estimation beyond proximity is restricted by the requirements for a large baseline, synchronized measurements and careful calibration.

Recent years have seen a significant breakthrough in Machine Learning (ML)-based monocular pose estimation of a known, non-cooperative spacecraft. Training of these ML models such as Convolutional Neural Networks (CNN) generally requires a large-scale dataset of images and pose labels. However, access to space is difficult and expensive, and a close-range rendezvous of two satellites is, in itself, a very rare occasion, making in-situ data collection and label annotation a challenging task for spaceborne applications. In response, the aerospace community has adopted a methodology whereby synthetic images are generated with rendering tools and gaming engines such as OpenGL and Unreal Engine in order to train the neural networks \citep{sharma_2020_taes_spn, proenca_2020_icra_urso, musallam_2021_icipc_spark, park_2024_scitech_spe3r}. However, synthetic images have inherently dissimilar visual features compared to real spaceborne images characterized by low Signal-to-Noise Ratio (SNR) and harsh illumination conditions. In addition, the degradation of vision cameras and target RSOs due to operations in the space environment is difficult to model prior to the mission and through computer graphics. This results in the problem known as \emph{domain gap}, which manifests as a drop in performance when the models are tested on the Out-Of-Distribution (OOD) data sampled from a statistical distribution different from that of the training data \citep{bendavid_2007_nips_domainadapt, peng_2018_cvprw_visda}. In case the models are trained with synthetic images and tested on real-life images, it is also known as \emph{reality gap} \citep{tremblay_2018_corl_dope} or \emph{sim2real gap} \citep{kadian_2020_ral_sim2real} primarily in the robotics literature.

\begin{figure}[t]
    \centering
    \includegraphics[width=\textwidth]{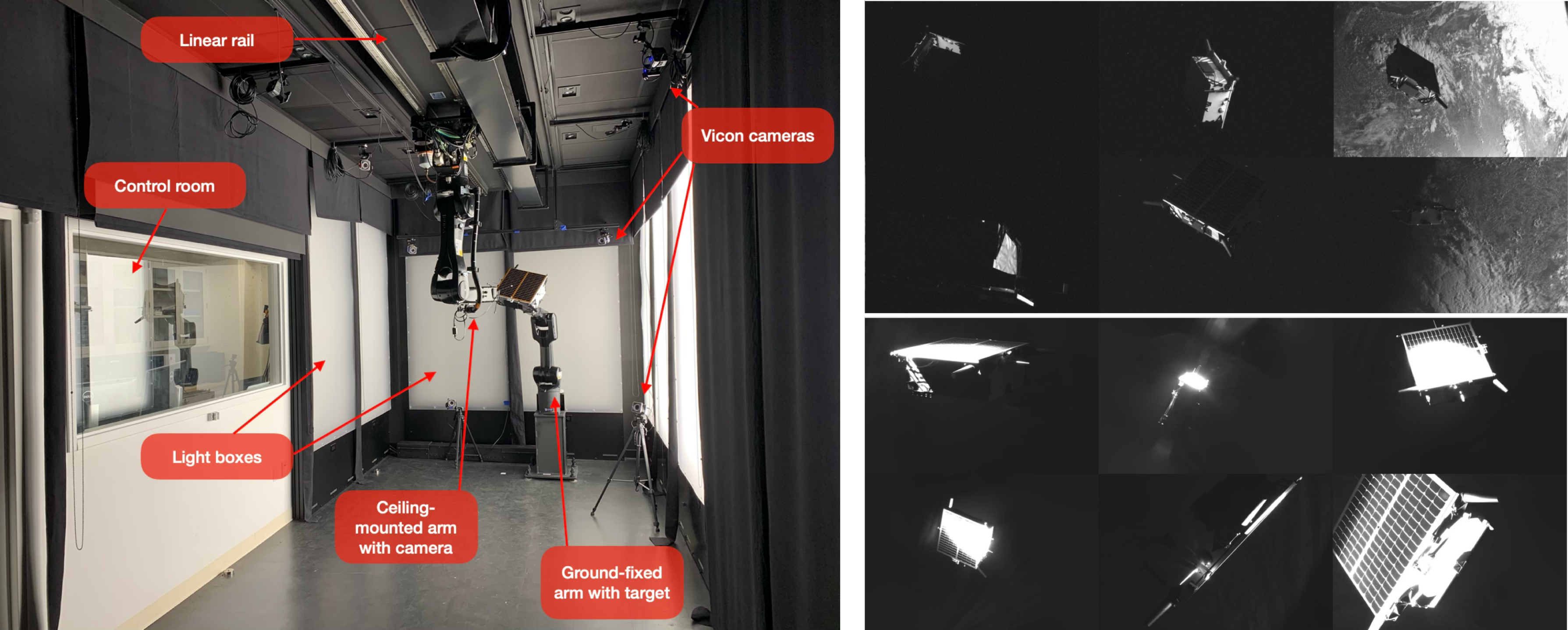}
    \caption{(\emph{left}) TRON simulation room and its components. Figure from \citet{park_2021_aas_tron}. (\emph{right}) Example HIL images from the $\lightbox$ (\emph{top}) and $\sunlamp$ domains (\emph{bottom}) of the SPEED+ dataset \citep{park_2022_aero_speedplus}.}
    \label{fig:01:tron_HIL_images}
\end{figure}

The inaccessibility of space poses a particularly challenging logistics problem to not only overcoming the domain gap but also verifying it during the stringent pre-flight verification processes typically required of space missions. The most prominent approach taken by the aerospace community is to utilize a robotic testbed to simulate the space environment on ground. The idea is to physically stimulate the vision-based sensors using a mockup model of the target placed in a facility simulating the high-fidelity space-like illumination conditions. A well-calibrated testbed can then be used to re-create various RPOD scenarios, generating pose-annotated images at scale with minimal human intervention for the purpose of validating a NN’s robustness across the sim2real gap. One such example is the Testbed for Rendezvous and Optical Navigation (TRON) facility at Stanford's Space Rendezvous Laboratory (SLAB)\footnote{\url{https://slab.stanford.edu/}} \citep{park_2021_aas_tron}, which was used to create the so-called Hardware-In-the-Loop (HIL) images that constitute datasets such as SPEED+ \citep{park_2022_aero_speedplus} and SHIRT \citep{park_2023_jgcd_spnukf}. \Cref{fig:01:tron_HIL_images} shows the TRON facility and a few example images from the two HIL domains of the SPEED+ dataset---$\lightbox$ and $\sunlamp$. In contrast to SPEED+, which contains static images at random poses, the SHIRT dataset includes sequential $\lightbox$ images obtained during dynamic RPOD scenarios. This allows the testing of navigation filters with ML algorithms in the loop. Overall, these HIL images can then be used as on-ground surrogates of otherwise unavailable spaceborne images for the evaluation of NN and the navigation algorithm robustness across domain gaps. 

SPEED+ was used as the core dataset for the second Satellite Pose Estimation Competition (SPEC2021)\footnote{\url{https://kelvins.esa.int/pose-estimation-2021/}} co-organized by SLAB and the Advanced Concepts Team (ACT) of the European Space Agency (ESA). From October 2021 to March 2022, participants were tasked to predict poses on SPEED+ HIL images while having access only to the labeled $\synthetic$ and unlabeled HIL domain images \citep{park_2023_acta_spec2021}. The submitted pose predictions were then evaluated based on a combination of translation and orientation errors separately for the \texttt{lightbox} and \texttt{sunlamp} domains. Finally, after the conclusion of the official competition, a separate post-mortem challenge was held until the end of 2022 in order to attract more innovative solutions for bridging the domain gap in spaceborne imagery.

Given such tools and datasets, the authors have explored a strategy to fully close the domain gap between synthetic and spaceborne images. The strategy consists of three steps. The first is to train a NN model using only synthetic images to be as robust as possible to the unknown spaceborne images \citep{park_2023_asr_spnv2}. The robustness is evaluated on ground using HIL domain images that are excluded from the training. Then, the NN is integrated as a measurement module into an Adaptive Unscented Kalman Filter (AUKF), which tracks the pose of the target \citep{park_2023_jgcd_spnukf}. While it was shown that a well-designed AUKF allows robust tracking of the target's pose despite the domain gap suffered by its NN measurement model, there is still a remaining gap between the HIL imagery used for on-ground evaluation and the spaceborne flight images. The primary reason is that HIL images use an inexpensive mockup model of the target spacecraft which has different material and surface properties compared to the real satellite. Therefore, \citet{park_2024_icra_ost} proposed to additionally fine-tune the NN weights using the flight images collected during in-space RPOD. The fine-tuning is done online in a supervised manner whereby the pose pseudo-labels are obtained from the most up-to-date state estimates of the onboard AUKF. This work has shown that Online Supervised Training (OST) can fully close the domain gap in the orientation prediction even when using a sub-optimally trained NN. 

Despite the success of OST, there are benefits to further maximizing the robustness of an onboard NN during the offline training on synthetic images. The first is that OST requires that the onboard ML-in-the-loop navigation filter be able to first converge to steady state so that accurate pose pseudo-labels can be generated from the state estimates. The probability of failure to converge is minimized as the OOD robustness of the onboard NN is maximized prior to deployment. The second is that training a NN onboard the limited computing environment of satellite avionics could be expensive both in terms of the computational effort and power consumption. This could pose a logistical challenge to schedule the NN training amidst the nominal Guidance, Navigation and Control (GN\&C) operations. These two reasons motivate
\begin{enumerate}
    \item a training mechanism that renders the NN as robust as possible across the domain gap from the offline training alone, which effectively minimizes the required number of OST steps performed in space, and
    \item a NN design that is computationally efficient for both inference and online training while maintaining the maximum OOD robustness.
\end{enumerate}
These two objectives---OOD robustness and computational efficiency---are at odds, since it is well known that larger NNs with higher capacity tend to perform better on both in-distribution and OOD data \mbox{\citep{tan_2019_icml_efficientnet, hendrycks_2021_iccv_manyfaces}}.

In response to the above challenge, the main contribution of this work is Spacecraft Pose Network v3 (SPNv3), a NN model based on Vision Transformer (ViT) \citep{dosovitskiy_2021_iclr_vit} for monocular pose estimation of a non-cooperative spacecraft. SPNv3 is designed and trained with onboard computational efficiency and robustness across the sim2real gap as top priorities. Extensive ablation studies of SPNv3 on the SPEED+ dataset demonstrate that ViT-based models without convolution operations dominate on the computational efficiency front, and that their OOD robustness on HIL domain images can be improved by employing extensive data augmentation, enhanced transfer learning and increased input image resolution with minimal gain of computational overhead for inference. A comprehensive set of experiments reveals that the proposed SPNv3 can achieve state-of-the-art performances on the HIL domain images of the SPEED+ dataset while training exclusively on its $\synthetic$ images. Most importantly, a mid-size variant of SPNv3 would rank the first place on the $\lightbox$ category of SPEC2021 \emph{without} any adversarial training or unsupervised domain adaptation, the methods which all winners employed by directly including the unlabeled test domain images into the training process. Such state-of-the-art robustness is achieved while taking no more than 40 ms per inference on an NVIDIA Jetson Nano 4GB \citep{nvidia_jetson_nano}, a representative, restricted compute environment similar to GPU-powered systems on satellite missions \citep{aitech_S_A1760, lajument_jetson}.

This paper is organized as follows: \Cref{sec:02_related} provides a brief overview of literature on ML-based monocular pose estimation and existing methods to bridge domain gap. \Cref{sec:03_req} then goes over three key requirements that must be met by spaceborne ML models in addition to OOD robustness. Two of them drive various design and training options for SPNv3 introduced in \cref{sec:04_searching} after briefly explaining the adopted pose estimation pipeline and assumptions in \cref{sec:035_scenario}. \Cref{sec:05_exp_ood_latency} shows extensive experiments on the SPEED+ dataset to identify key elements of the design and training of SPNv3 that contribute the most to its robustness on HIL domain images and computational efficiency. The paper ends with conclusions and future works in \Cref{sec:07_conclusion}.

\section{Related Work}
\label{sec:02_related}

\subsection{ML Approaches to Spacecraft Pose Estimation}
\label{sec:02_related:pose_est}

One of the first ML-based approaches to pose estimation of a known target spacecraft was Spacecraft Pose Network (SPN) \mbox{\citep{sharma_2017_icssa_pose, sharma_2018_aero_pose, sharma_2020_taes_spn}} which performs 1) relative attitude determination via a hybrid approach of attitude classification and regression and 2) translation estimation by exploiting the perspective transformation and geometric constraints. Since it is impractical to collect a pose-annotated image dataset from space, the same work introduced the Spacecraft PosE Estimation Dataset (SPEED) \citep{sharma_2019_speed} which consists of 15,300 images of the Tango spacecraft from the PRISMA mission \citep{damico_2013_prisma}. Specifically, the dataset comprises 1) 15,000 synthetic images rendered with OpenGL-based Optical Stimulator (OS) camera emulator software \citep{beierle_2019_jsr_os, sharma_2018_aero_pose} of SLAB multi-Satellite Software Simulator ($S^3$) \citep{giralo_2018_ion_gnss} and 2) 300 \emph{real} images of a mockup of the same target captured from the TRON facility at SLAB. The dataset was made publicly available as part of the first Satellite Pose Estimation Challenge (SPEC2019)\footnote{\url{https://kelvins.esa.int/satellite-pose-estimation-challenge/}} \citep{kisantal_2020_taes_spec} co-hosted by SLAB and ESA.

\begin{figure}[!t]
	\includegraphics[width=0.7\textwidth]{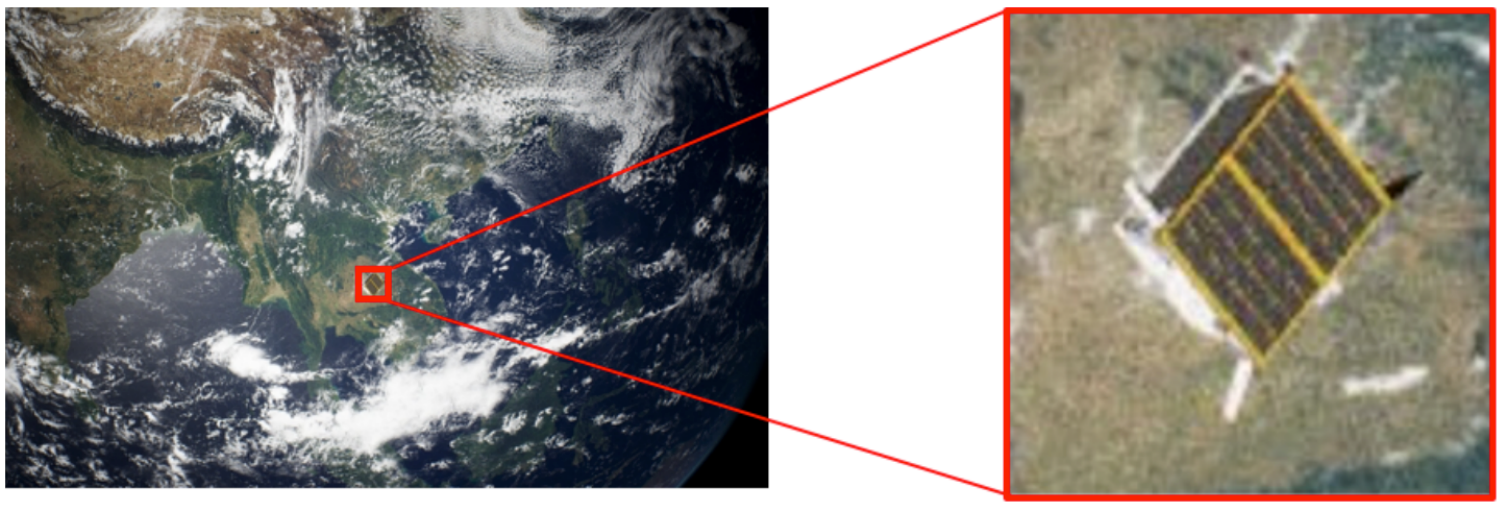}
	\centering
	\caption{Visualization of an image cropped around the far-away target spacecraft.}
	\label{fig:02:image_crop}
\end{figure}

Many top-performing entries of SPEC2019 have adopted diverse CNN architectures and pose estimation strategies, such as probabilistic orientation estimation via soft classification \citep{proenca_2020_icra_urso} or estimation of 2D pixel coordinates of the designated keypoints on the spacecraft surface \citep{park_2019_aas_krn, chen_2019_iccvw_pose}. Specifically, \citet{chen_2019_iccvw_pose} and \citet{park_2019_aas_krn}, who respectively ranked first and fourth places in the challenge, independently proposed a three-stage architecture: 1) an object detection CNN which is used to identify the Region-of-Interest (RoI) around the target, 2) a pose estimation CNN which takes in an image cropped around the detected RoI and outputs the 2D keypoint locations, and 3) a P$n$P module which solves for the full 6D pose based on the known correspondence of detected keypoint 2D locations and 3D model coordinates. Notably, \citet{park_2019_aas_krn} directly regresses $(x, y)$ coordinates of the keypoints, whereas \citet{chen_2019_iccvw_pose} outputs a set of 2D heatmaps whose peaks correspond to the locations of the keypoints. They also showed that cropping the input image around the target using a detected bounding box is crucial to performing pose estimation of a far-away target. For example, the original resolution of the SPEED images is 1920 $\times$ 1200, while most CNNs for ImageNet \citep{russakovsky_2015_ijcv_imagenet} classification expect 224 $\times$ 224 inputs. Cropping around the target prior to such dramatic down-scaling helps preserve much of the detailed target features that would otherwise be lost due to large inter-spacecraft separation as visualized in \cref{fig:02:image_crop}. Many CNN models that followed SPEC2019 also adopt similar strategies \citep{garcia_2021_cvprw_lspnet, hu_2021_cvpr_swisscube, kaki_2023_jais_pose, musallam_2021_icipc_spark, pasqualetto_2021_acta_cnn, lotti_2023_jsr_tpu}. Readers are referred to \citet{pasqualetto_2019_pas_review} for a more comprehensive review of monocular spacecraft pose estimation using both conventional and deep learning-based methods.

In addition to SPEED, a number of datasets for spacecraft pose estimation have been published and made publicly available in the literature. For example, \citet{proenca_2020_icra_urso}, who also ranked third place in SPEC2019, published the Unreal Rendered Spacecraft On-orbit (URSO) dataset, which uses Unreal Engine 4 to render synthetic images of the Soyuz and Dragon spacecraft. \citet{kaki_2023_jais_pose} renders synthetic images of the Cygnus spacecraft using Blender and its Cycles rendering engine. \citet{dung_2021_cvprw_dataset} also renders about 3,000 synthetic images of different satellites for various computer vision tasks such as bounding box prediction and satellite foreground and parts segmentation. Other authors also created their own synthetic datasets, such as those of the Envisat spacecraft rendered with Cinema 4D \citep{pasqualetto_2021_acta_cnn}, the SwissCube dataset for spacecraft pose estimation from wide-depth-range images using the Mitsuba 2 renderer \citep{hu_2021_cvpr_swisscube}, the SPARK \citep{musallam_2021_icipc_spark} dataset which contains images of 11 different spacecraft rendered with the Unity3D game engine, and Synthetic-Minerva II2 \citep{price_2021_cvprw_minerva} which renders images of the Minerva-II2 rover from the Hayabusa2 mission \citep{tsuda_2013_acta_hayabusa} using SolidWorks' Photoview 360 renderer.

\subsection{Algorithms to Bridge Domain Gap}
\label{sec:02_related:domaingap}

Approaches to tackling domain gaps can be put into two categories: domain adaptation and domain randomization. Specifically, Unsupervised Domain Adaptation (UDA) methods aim to close the gap by directly incorporating the \emph{unlabeled} target-domain images into the training phase \citep{bendavid_2007_nips_domainadapt, bousmalis_2017_cvpr_adapt, ganin_2016_jmlr_dann, long_2017_icml_jmmd, peng_2018_cvprw_visda, sun_2016_eccv_deepcoral, tsai_2018_cvpr_adapt, tzeng_2017_cvpr_adda, yan_2017_cvpr_wmmd}. One primary approach is to align the source and the target data distributions in the latent feature space by minimizing a certain divergence criterion representing the domain discrepancy, such as maximum mean discrepancy \citep{gretton_2012_jmlr_kernel, long_2017_icml_jmmd, wang_2018_acm_adapt, yan_2017_cvpr_wmmd}, $\mathcal{H}$-divergence \citep{bendavid_2007_nips_domainadapt, ganin_2016_jmlr_dann}, correlation via statistical moments \citep{peng_2019_iccv_moment, sun_2016_eccv_deepcoral} or Wasserstein distance \citep{damodaran_2018_eccv_deepjdot, redko_2017_mlkd_ot_adapt, shen_2018_aaai_wasserstein}. Another approach employs adversarial training \citep{ganin_2016_jmlr_dann, hoffman_2018_icml_cycada, tzeng_2017_cvpr_adda} so that the predicted features become indistinguishable for both source and target domain inputs from the perspective of an auxiliary domain discriminator network. Note that the aforementioned approaches all aim to promote learning invariant representations by aligning both source and target domain data in the feature space.

One major shortcoming of UDA is that it requires simultaneous availability of the labeled source and unlabeled target-domain data. This assumption violates the operational constraints of space missions, as spaceborne images of the target do not become available until on-orbit rendezvous. Moreover, the processor and memory onboard the satellites are extremely limited for the full training session using large-scale synthetic source images and spaceborne target images collected during RPOD. Therefore, the conventional UDA approaches cannot be used for realistic space mission scenarios. 

As opposed to UDA, domain randomization instead aims to randomize various aspects of the training images such that OOD test images would appear as another randomized instance of the training set \citep{prakash_2019_icra_sdr, sadeghi_2017_rss_cad2rl, tobin_2017_iros_domainrandom, yue_2019_iccv_random, zakharov_2019_iccv_deceptionnet}. For example, \citet{sadeghi_2017_rss_cad2rl} and \citet{tobin_2017_iros_domainrandom} randomize lighting conditions, object textures and placements at the rendering stage to train domain randomized reinforcement learning models. \citet{jackson_2019_iccvw_styleaug} proposes style augmentation, which randomizes the image texture \citep{geirhos_2019_iclr_imagenetbias} via neural style transfer \citep{gatys_2016_cvpr_nst}. While domain randomization does not require target domain images during the training phase, it also makes it difficult to provide any assurance that the CNN trained with domain randomization will work well on the target spaceborne images.

\subsubsection{Source-Free Domain Adaptation}
\label{sec:02_related:domaingap:sfda}

A different approach conducive to the operational constraints of space missions is \emph{source-free} domain adaptation, which does not require the availability of large-scale training images during the adaptation phase. Existing source-free algorithms leverage generative models for feature alignment \citep{kurmi_2021_wacv_domain_impress, li_2020_cvpr_modeladapt, yeh_2021_wacv_sofa} or pseudo-labeling and information maximization \citep{liang_2020_icml_hypo_transfer}. TENT \citep{wang_2021_iclr_tent} performs entropy minimization while updating only the affine parameters of the Batch Normalization (BN) layers \citep{ioffe_2015_icml_batchnorm}. However, methods based on entropy minimization require optimizations to be performed on batches of images in order to avoid trivial solutions, which could become computationally expensive on satellite avionics.

On the other hand, Test-Time Training (TTT) \citep{sun_2020_icml_ttt, liu_2021_nips_ttt++} trains on a secondary Self-Supervised Learning (SSL) task from a shared feature encoder. During test time, the encoder is trained on SSL tasks, such as rotation prediction \citep{gidaris_2018_iclr_rot_pred} or image colorization \citep{larsson_2016_eccv_image_color}. However, TTT generally requires a hand-designed task or a large batch of negative sample pairs (e.g., contrastive learning \citep{chen_2020_icml_contrastive}). Finally, \citet{lu_2022_iros_slam_self_train} self-trains an object pose estimator CNN using pseudo-labels from its own predictions. In order to improve the accuracy of pseudo-labels and mitigate outliers, they take a SLAM-based approach and solve the Pose Graph Optimization (PGO) problem to enhance the consistency of pseudo-labels across different images. However, PGO is an offline problem solving for a set of multiple poses satisfying the motion constraints, and collecting many images until pose pseudo-labels can be obtained via PGO could become computationally expensive on satellite avionics.

\subsubsection{Spaceborne Applications}
\label{sec:02_related:domaingap:spaceborne}

The approaches to bridging the domain gap in spaceborne applications have been explored recently after the introduction of SPEED+ \citep{park_2022_aero_speedplus} and SPEC2021 \citep{park_2023_acta_spec2021}. Two winners of the competition on respective HIL domain categories have both employed a generative-adversarial training procedure \citep{goodfellow_2014_nips_gan}, introducing a discriminator network to the NN outputs to classify whether the predictions are made on the $\synthetic$ or HIL domain images. The winner of the $\sunlamp$ category further employed pseudo-labeling and self-training on the unlabeled HIL domain images \citep{wang_2023_taes_lava1302}. \citet{perez_villar_2022_eccv_vpu}, who placed second place on both categories, also employed UDA and pseudo-labeling of the HIL domain images. 

On the other hand, SPNv2 \citep{park_2023_asr_spnv2}, one of the baseline models developed by the authors, designed a multi-scale, multi-task learning CNN architecture and trained with extensive data augmentations on the SPEED+ synthetic images only. The largest variant of SPNv2 would have ranked 3rd place in $\lightbox$ and 6th in $\sunlamp$ categories, respectively, without accessing the HIL domain images during training at all. Finally, EagerNet \citep{ulmer_2023_iros_eagernet} performs dense predictions of pixel-wise object coordinates as opposed to heatmap predictions that many other methods adopted. They additionally predict the errors of predicted model coordinates, which results in multiple pose hypotheses that are further refined using a probabilistic refinement. EagerNet trained with extensive data augmentation, including those specifically designed to target the HIL domain imagery, achieves state-of-the-art robustness without accessing the HIL domains, winning both HIL categories in the post-mortem competition.

As shown later, SPNv3 is able to achieve comparable robustness relative to EagerNet without access to the HIL domains as well. However, SPNv3 is a much simpler architecture that outputs heatmaps about known keypoints on the target surfaces, which can be provided directly to the onboard navigation filter \citep{park_2023_jgcd_spnukf} or be used to solve for the optimal pose solution via P$n$P \citep{lepetit_2008_ijcv_epnp}.

\section{Requirements of Spaceborne ML Models} 
\label{sec:03_req}

Before introducing SPNv3, this section provides a brief overview of various requirements that must be met by not only SPNv3, but also ML models that are intended to run in space onboard satellite avionics. Some of these requirements, in addition to the OOD robustness across the sim2real gap, drive the design choices of SPNv3 in the next section.

\subsection{Computationally Efficiency}
\label{sec:03_req:efficiency}

The first requirement of a spaceborne ML model operating on satellite avionics is its computational efficiency. This is an obvious requirement for real-time operation of NNs in space whose runtime should ideally be within the update frequency of its overarching GN\&C system. For example, \citet{roscoe_2018_acta_cpod} reports that the CubeSat Proximity Operation Demonstration (CPOD) mission launched in 2022 ran its RPO GN\&C system at 0.5 Hz frequency with its star sensor and IMU data processed at up to 10 Hz for its Attitude Determination and Control System (ADCS) as part of an Extended Kalman Filter (EKF). However, the image processing algorithms often exceeded the onboard filter measurement update cycles, requiring a robust filter design that can handle out-of-sync and latent measurements. Taking the CPOD mission as a reference, this means the measurement processing of a NN, which includes not only its forward pass but also pre- and post-processing of its inputs and outputs, must be completed well within two seconds on a representative computing hardware with limited processing capabilities. Achieving this could be facilitated by the presence of an onboard GPU that can perform the NN inference and any other parallelizable operations, allowing the inference to be executed asynchronously from other GN\&C operations running on Central Processing Units (CPU).

As explained in \cref{sec:01_intro}, it may be desirable to perform additional online training of NNs in space to fully close the domain gap by directly incorporating the flight images that only become available during RPOD. In this case, the NN must also be computationally efficient to run not only forward but also backward gradient propagation. Unlike inference, the training most certainly requires a GPU as it is computationally expensive to track gradients across the entire NN. However, \mbox{\citet{park_2024_icra_ost}} has shown that performing one training iteration whenever there has been a substantial change in the spacecraft pose is sufficient to improve the NN's OOD robustness. This not only renders onboard training computationally feasible but also prevents overfitting the NN to the specific scenery (e.g., Earth background) and view of the target. Therefore, under this strategy, as long as there is a GPU onboard, the training latency is likely to be less of a computational bottleneck than inference, which should be running in real-time. Readers are referred to \mbox{\citet{park_2024_icra_ost}} for more details on online supervised training of NN onboard spacecraft avionics.

\subsubsection{Remark on GPU Flight Heritage}
\label{sec:03_req:efficiency:gpu}

Even nowadays, GPUs are still rarely found onboard spacecraft, and only in short-term missions which typically are not concerned with long-term radiation effects such as Total Ionising Dose (TID), or as part of a larger spacecraft which provides better protection of the onboard computing systems. Recent state-of-the-art report on small spacecraft technology by NASA has tabulated a number of GPU-based avionics \citep[Table 8.1, \S 8.3]{nasa_soa_smallsat_2023}, but many Commercially Off-The-Shelf (COTS) systems such as NVIDIA Jetson series are still not flight-proven or are radiation tolerant only against Single Event Effects (SEE) (e.g., bit flips). 

However, there is an increasing interest in deploying GPUs into deep space or long-term missions in LEO with existing flight heritage \citep{lajument_jetson}. For example, Aitech Systems, Inc., a company that manufactures rugged computers for military and aerospace applications, utilizes the NVIDIA Jetson TX2i System-on-Module (SoM) in their S-A1760 system \citep{aitech_S_A1760} which is designed for spacecraft and small satellites and has a flight heritage in LEO as part of a larger payload system \citep{aitech_S_A1760_flight}. The startup company Aethero\footnote{\url{https://www.aethero.com/}} is also developing radiation-hardened edge computers for on-orbit data processing and autonomous decision-making in space. Its ECM-NxN utilizes an NVIDIA Jetson Orin processor, considered the best GPU edge processor available. While the ECM-NxN module has not directly flown in space, Aethero's use of this processor in their space computer indicates potential applications in space exploration. Finally, the Ingenuity Mars Helicopter used a GPU-equipped Qualcomm\textsuperscript{\textregistered} Snapdragon\textsuperscript{\texttrademark} 801 processor with great results, even though it operated in a ``planetary'' environment \citep{ingenuity_qualcomm}. These are a few examples of global efforts to deploy edge GPU systems to space, and there is no doubt that more and more powerful space-grade GPUs will become commercially available to endure the harsh space environment for a prolonged period of time.

\subsection{Number of Parameters}
\label{sec:03_req:num_params}

The number of NN parameters is closely related to the computational efficiency since more weights translate to more operations to be performed for both training and inference. However, the NN size is also tightly connected to the power consumption. For instance, the seminal work of \citet{han_2015_nips_pruning} in 2015 identified that, under the 45nm Complementary Metal-Oxide Semiconductor (CMOS) technology, 32-bit memory access to a Dynamic Random Access Memory (DRAM) consumes three orders of magnitude more power than accessing a Static Random Access Memory (SRAM) cache or performing a 32-bit floating point arithmetic. Considering miniature satellite systems such as CubeSats with limited power generation capability, it would be favorable to minimize the NN size so that the entire network can be hosted on a local SRAM, which is limited in memory size (typically up to 20MB for radiation hardened SRAM \citep[Table 8.2, \S 8.3.4]{nasa_soa_smallsat_2023}).

The reduction in NN memory footprint can be achieved not just by minimizing the number of learnable parameters but also by quantizing the weights to lower-precision values such as IEEE half-precision floating point (FP16) \citep{micikevicius_2018_iclr_mixedprecision} and 8-bit integer (INT8) \citep{krishnamoorthi_2018_quant}. More recent innovations in quantized training and inference include brain floating point (\texttt{bfloat16}) \citep{google_bfloat16} for half-precision, 8-bit floating point (FP8) \citep{sun_2019_nips_fp8} and 4-bit floating point (FP4) \citep{sun_2020_nips_fp4}, and even binary CNN whose weights and activations are constrained to $\{-1, +1\}$ \citep{lin_2017_nips_binary_cnn}. Unfortunately, working with quantized weights often requires care to prevent over/underflowing of floating point values during gradient backpropagation and to minimize the loss of accuracy due to reduced bit resolution of individual weights. Moreover, many quantized floating point formats require specific hardware support to realize actual computational gain.

While power consumption of operating NNs on satellite avionics is an important engineering topic, this work does not immerse into the subject as realizing NN quantization is often an issue of software and hardware support on a specific system. Instead, it focuses on more general approaches to minimizing the total number of parameters while realizing the best OOD robustness on the SPEED+ dataset.

\subsection{Batch-Agnostic Architecture}
\label{sec:03_req:batch_agnostic}

If any online training is to be done, it is desirable to have a batch-agnostic NN architecture, i.e., it does not contain any Batch Normalization (BN) layers \citep{ioffe_2015_icml_batchnorm}. The core of normalization (\norm) is to prevent internal covariate shifts of the feature maps and stabilize the training process. This is done by normalizing each feature map with a mean and a variance, for an $i$-th layer feature map,
\begin{align} \label{eqn:03:batch_agnostic:norm_op}
    \hat{\bm{X}}_i = \norm(\bm{X}_i) \triangleq \frac{\bm{X}_i - \mathbb{E}[\bm{X}]}{\sqrt{\text{Var}[\bm{X}_i]}}
\end{align}
where $\bm{X}_i$ is the feature map at the $i$-th layer. Unlike other methods, BN aims to compute the \emph{batch-wise} feature map statistics across the entire training set. However, computing such statistics across the entire training domain is practically infeasible given a limited dataset. Therefore, the BN layer statistics are instead computed across mini-batches, and the approximate ``global'' statistics are updated using a running average as training progresses. Then, during inference, these running averages of the mean and variance are fixed and used for normalization.

The problem is that these approximations of the training data distribution in BN layers are simply incompatible with those of the test data that are drawn from a different distribution. Therefore, in this work, two methods are considered in search of a robust and batch-agnostic pose estimation NN architecture. One is to simply replace the BN layers in existing pre-trained architectures with batch-agnostic $\norm$ layers such as Layer Normalization (LN) \citep{ba_2016_layernorm} and Group Normalization (GN) \citep{wu_2018_eccv_groupnorm}. The other is to design or adopt a NN architecture that is inherently without BN layers such as ViT \citep{dosovitskiy_2021_iclr_vit}.

\subsection{Summary}
\label{sec:03_req:summary}

In summary, in order to ensure that SPNv3 is suitable for real-time onboard inference and fine-tuning, it must meet the following objectives:
\begin{itemize}
    \item it must be able to process incoming images within a representative ADCS system frequency (e.g., 0.5 Hz)
    \item it must have a minimum number of learnable parameters in order to reduce the memory footage
    \item considering the possibility of onboard training and fine-tuning, it must have minimum dependence on NN training components that must operate on a batch of images (e.g., BN layers)
\end{itemize}
All the above must be accomplished while simultaneously maximizing the pose estimation accuracy on OOD spaceborne images.

\section{Pose Estimation Pipeline}
\label{sec:035_scenario}

\begin{figure}[!t]
	\includegraphics[width=\textwidth]{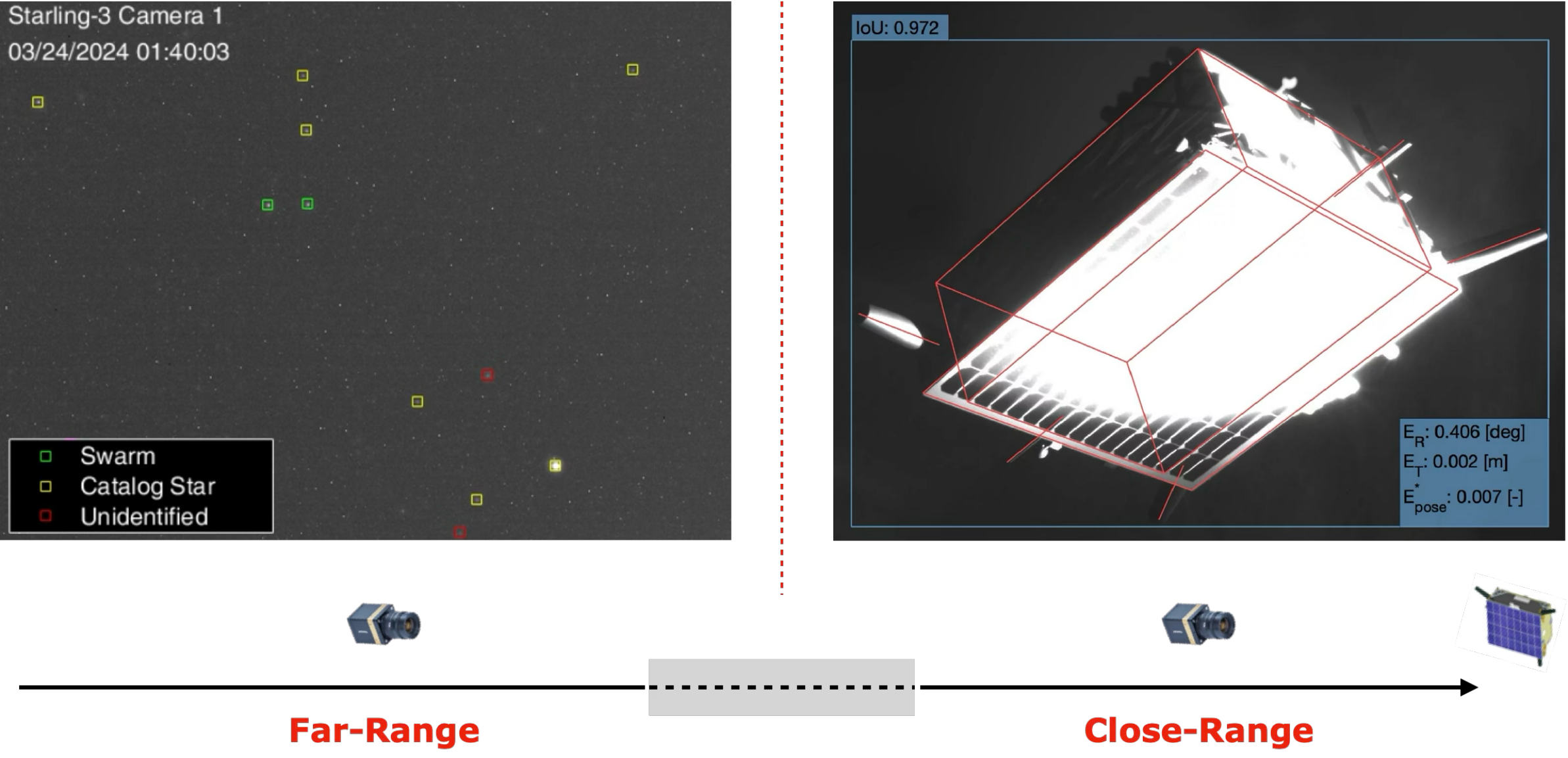}
	\centering
	\caption{Visualization of far-range AON \citep{kruger_2021_acta_aon} (\emph{left}) and close-range pose estimation (\emph{right}).}
	\label{fig:04:scenario:operation}
\end{figure}

In all vision-based RPOD scenarios, the servicer spacecraft begins tracking the non-cooperative target at kilometers of separations as shown in \cref{fig:04:scenario:operation}. For example, Angles-Only Navigation (AON) \citep{damico_2013_jgcd_aon, sullivan_2021_jgcd_aon, kruger_2021_acta_aon, kruger_2023_jsr_starfox} obtains bearing angle measurements of the target from a Narrow Field-Of-View (NFOV) camera, such as a star tracker, which allows tracking of the target's relative orbital state via nonlinear filtering. AON continues until the inter-spacecraft distance becomes small enough such that the target appears resolved in the camera view, at which point the pose estimation algorithm kicks in. Note that in general, the sensors are selected to ensure overlapping working distance to the target during the transition from far-range to close-range RPO, but it is not always possible based on the orbit and size of the target.

\begin{figure}[!t]
	\includegraphics[width=\textwidth]{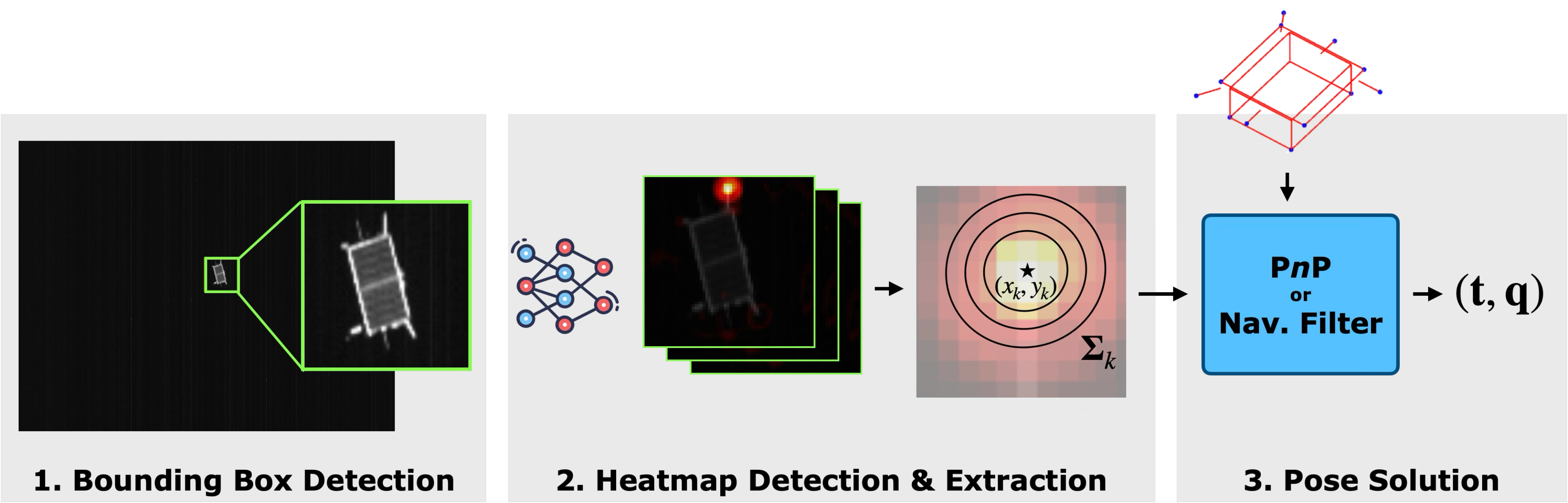}
	\centering
	\caption{Visualization of the pose estimation pipeline adopted in this work.}
	\label{fig:04:scenario:pipeline}
\end{figure}

Given the scenario, this work adopts the three-stage approach visualized in \cref{fig:04:scenario:pipeline} as the common mechanism of the ML-based pose estimation algorithm, which was first independently proposed by \citet{park_2019_aas_krn} and \citet{chen_2019_iccvw_pose} for SPEC2019 \citep{kisantal_2020_taes_spec}:
\begin{enumerate}
    \item The first stage detects the target within the image frame using the state estimates from AON or a separate object detection NN. The latter may become especially useful in the case of the ``lost-in-space'' scenarios in which the navigation filter must reset all its state estimates. The detected bounding box or Region-of-Interest (RoI) from these sources is used to crop the image around the target spacecraft, and the cropped image is resized to lower-resolution inputs expected by the second-stage pose estimation NN (e.g., 224 $\times$ 224).
    \item The second stage is the main pose estimation NN, which takes the cropped low-resolution images and outputs $K$ heatmaps centered at the 2D locations of known target surface keypoints. The advantage of dense predictions such as heatmaps is that their spread about the peak can be interpreted as uncertainties associated with each prediction \citep{pasqualetto_2021_acta_cnn}. Therefore, at each $k$-th heatmap, in addition to its mean location $(x_k, y_k)$, its spread is approximated as a covariance matrix $\bm{\Sigma}_k$ and returned.
    \item The final stage takes the 2D keypoint locations and, knowing the 3D coordinates of these keypoints in the body frame, solves P$n$P to compute the full 6D pose solution. The uncertainty covariances can also be incorporated into the uncertainty-aware P$n$P algorithm \citep{peng_2019_cvpr_pvnet, pasqualetto_2021_acta_cnn}, or the mean and covariance can both be provided as measurements to the navigation filter \citep{park_2023_jgcd_spnukf} which tracks the target's pose.
\end{enumerate}
The heatmap-based pose estimation NN architecture is the most prominent approach, having won the first places in both SPEC2019 \citep{kisantal_2020_taes_spec} and SPEC2021 \citep{park_2023_acta_spec2021}. Therefore, this work adopts the same heatmap detection architectures for NN. The remainder of this section explores various architectural designs and training algorithms to make the NN as robust as possible to OOD data.
\section{Searching for Next Spacecraft Pose Network (SPNv3)}
\label{sec:04_searching}

In this section, different NN architectures (\cref{sec:04_searching:arch}) and data augmentation techniques (\cref{sec:04_searching:augment}) for training a pose estimation NN are explained which are combined in various configurations to study and explore which aspects of NN architecture and training algorithm contribute the most to the OOD robustness and reduced latency of a pose estimation NN (Section \ref{sec:05_exp_ood_latency}). While SPN is the name of the first spacecraft pose estimation CNN introduced by \citet{sharma_2020_taes_spn}, it is also used throughout this work as a legacy name referring to all NN architectures designed or adopted for spacecraft pose estimation by SLAB. When referring to the first original SPN, it is always accompanied by the reference to the relevant work \citep{sharma_2020_taes_spn}.

\subsection{Pose Estimation Architectures}
\label{sec:04_searching:arch}

\begin{figure*}[!t]
	\includegraphics[width=1.0\textwidth]{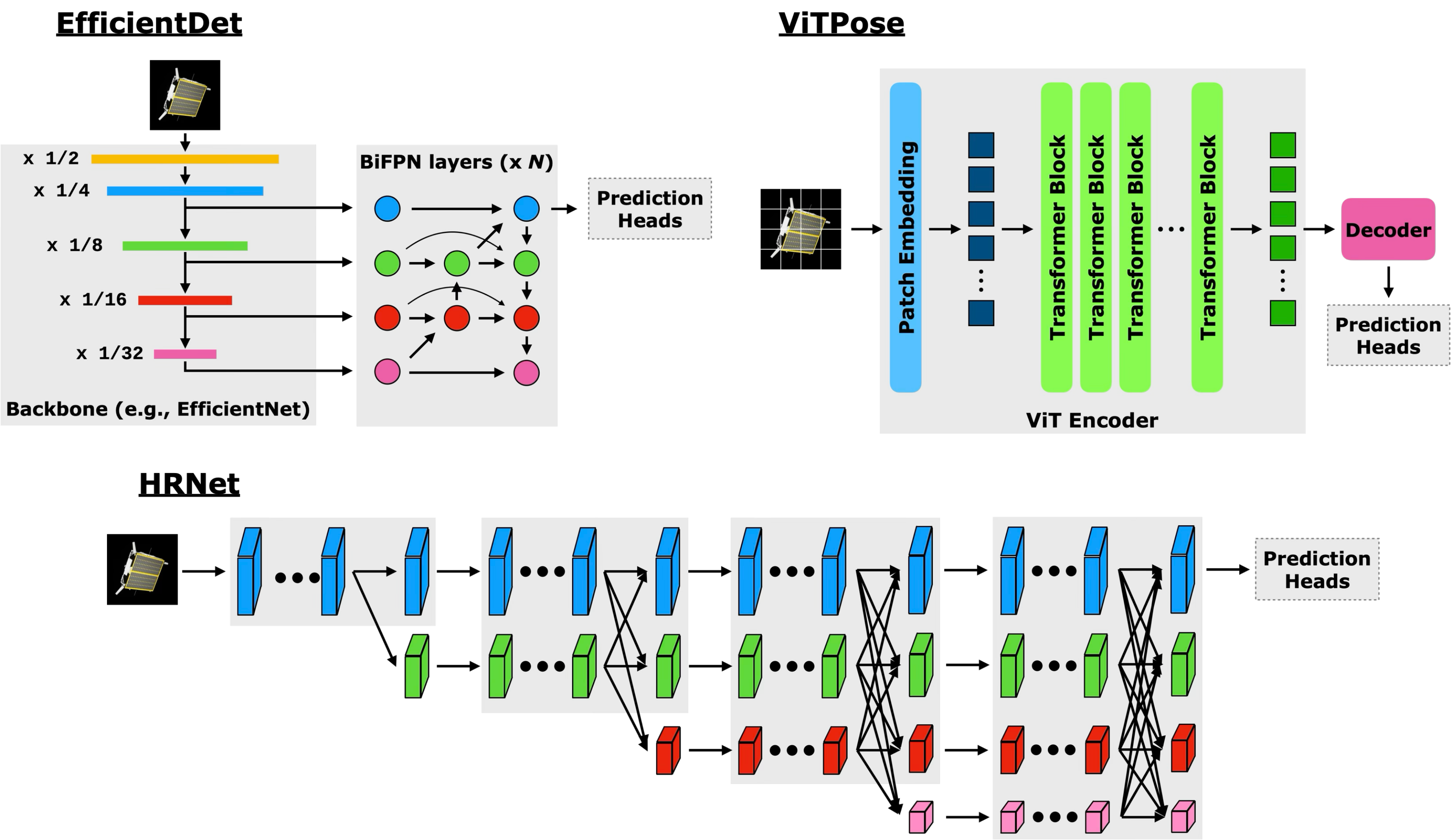}
	\centering
	\caption{Visualization of different pose estimation architectures: EfficientDet \citep{tan_2020_cvpr_efficientdet}, HRNet \citep{wang_2021_tpami_hrnet} and ViTPose \citep{xu_2022_nips_vitpose}. The architectures commonly consist of a backbone/encoder network and a heatmap prediction head.}
	\label{fig:04:pose_arch:archs}
\end{figure*}

This section explores three different heatmap detection architectures---EfficientDet \citep{tan_2020_cvpr_efficientdet}, HRNet \citep{wang_2021_tpami_hrnet} and ViTPose \citep{xu_2022_nips_vitpose}---visualized in \cref{fig:04:pose_arch:archs}. Note that this is not a comprehensive list of state-of-the-art pose estimation NN architectures. Nonetheless, they are chosen due to their simplicity and in order to keep the number of training sessions tractable while exploring vastly different architectural options.

\subsubsection{EfficientDet}

EfficientDet was originally proposed by \citet{tan_2020_cvpr_efficientdet} to enable computationally efficient yet high-performance object detection and semantic segmentation. EfficientDet builds upon the EfficientNet \citep{tan_2019_icml_efficientnet} backbone designed for the image classification task. At the heart of EfficientDet is the Bidirectional Feature Pyramid Network (BiFPN), a novel FPN architecture \citep{lin_2017_cvpr_fpn} that fuses and processes feature outputs from the backbone at multiple scales. By fusing features from low- to high-resolution and vice versa, BiFPN outperforms previous state-of-the-art FPN architectures on various benchmark object detection and semantic segmentation datasets. In the end, the feature outputs from BiFPN are processed by the task-specific prediction heads composed of convolution layers. For tasks that require multi-scale predictions (e.g., object detection), features from all resolutions are processed. In this work, the heatmap prediction head is attached only to the P2 level feature as shown in \cref{fig:04:pose_arch:archs}, where P$n$ indicates that the feature size is 1 / $2^n$ of the input image resolution. Only the P2 features (i.e., heatmap has 1/4 of the input resolution) are used since the accuracy of heatmap prediction heavily depends on the heatmap pixel resolution.

\subsubsection{HRNet}

High Resolution Network (HRNet) \citep{wang_2021_tpami_hrnet} is also designed for performing tasks that require multi-scale predictions. Recall that EfficientDet builds upon a CNN backbone that is typically designed for the image classification task, resulting in an architecture which continuously reduces the feature resolution all the way down to 1/32 of the original input. On the other hand, HRNet is designed to maintain high-resolution features throughout the entire forward propagation, outputting the multi-resolution features by the nature of its design. To facilitate learning at different scales, HRNet first processes the input image to 1/4 resolution, then gradually adds lower-resolution features in parallel, continuously fusing information across different scales as visualized in \cref{fig:04:pose_arch:archs}.

HRNet has won SPEC2019 \citep{chen_2019_iccvw_pose, kisantal_2020_taes_spec} and the $\lightbox$ category of SPEC2021 \citep{park_2023_acta_spec2021}, so it is considered as a candidate benchmark in this paper. Similar to EfficientDet, only the highest resolution P2 level features are processed for heatmap outputs.

\subsubsection{ViTPose}

Compared to EfficientDet and HRNet which are CNN architectures based on convolution operations, ViTPose \citep{xu_2022_nips_vitpose} leverages the ViT backbone \citep{dosovitskiy_2021_iclr_vit} which divides the input image into $P \times P$ patches then processes them in parallel through transformer blocks. ViTPose groups the output token embeddings according to their original spatial position in the input image. The resulting features have the resolution of $H/P \times W/P$, where $(H, W)$ is the input image size. For example, the output features will have 1/16 resolution of the original image for $P = 16$. ViTPose processes them through 2 transposed convolution layers in the prediction head which upsamples the input feature to 1/4 resolution heatmap outputs, same as EfficientDet and HRNet.

\subsection{Extensive Data Augmentation}
\label{sec:04_searching:augment}

Data augmentation is vital to training NNs that generalize well beyond their training data distribution and to preventing overfitting. As basic sets of augmentation, SPN heavily leverages the Albumentations library \citep{buslaev_2020_albumentations} for various image processing operations. For each training sample, $N$ augmentation operations are randomly chosen from the set of designated operations in a manner similar to RandAugment \citep{cubuk_2020_nips_randaugment}. Specifically, while RandAugment utilizes a single hyperparameter to control the constant level of magnitude of these operations (e.g., standard deviation of white Gaussian noise), this work simply picks each operation with a random magnitude within some pre-defined window. Then, these $N$ operations are applied to the original image in sequence. The complete list of employed augmentations is provided in \cref{tab:04:searching:augmentation:albumentations} along with their equivalent Albumentations commands. All commands are run with the default range of magnitudes.

\begin{table}[t]
\caption{List of data augmentations and equivalent commands in Albumentations v1.3.1 \citep{buslaev_2020_albumentations}.}
\label{tab:04:searching:augmentation:albumentations}
\centering
\small
\tabcolsep=0.075cm
\begin{tabular}{ll}
\toprule
Augmentations & Commands \\
\midrule
Brightness \& Contrast & \texttt{RandomBrightnessContrast} \\
Blur & \texttt{OneOf(GaussianBlur, MedianBlur, Defocus)} \\
Noise & \texttt{OneOf(GaussNoise, ISONoise, MultiplicativeNoise)} \\
Bit Reduction & \texttt{Posterize} \\
Sharpen & \texttt{Sharpen} \\
Pixel Inversion & \texttt{Solarize} \\ 
Spatter & \texttt{Spatter} \\
Histogram Equalization & \texttt{Equalize} \\
Gamma Correction & \texttt{RandomGamma} \\
Solar Flare\textsuperscript{\textdagger} & \texttt{RandomSunFlare} \\
\bottomrule
\multicolumn{2}{l}{\textsuperscript{\textdagger}Modified so that the solar flare only appears within the ground-truth bounding box.}
\end{tabular}
\end{table}

\begin{figure}[!t]
    \includegraphics[width=1.0\textwidth]{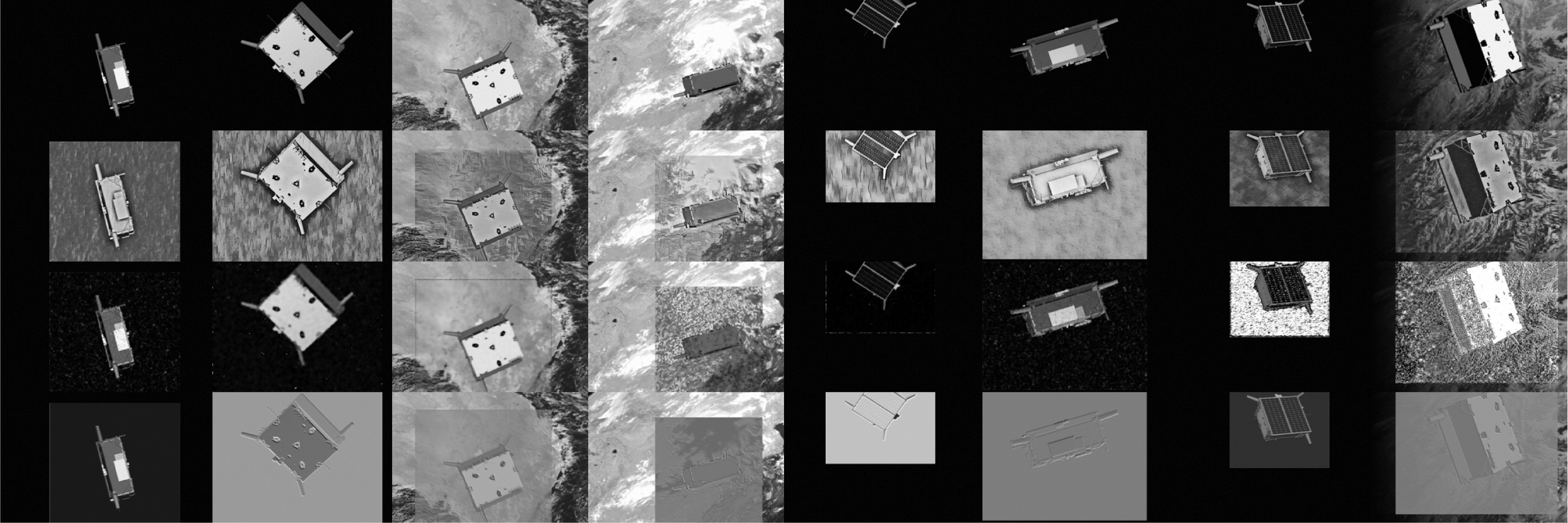}
    \centering
    \caption{Image augmentations. From top to bottom: SPEED+ $\synthetic$ images; StyleAugment \citep{jackson_2019_iccvw_styleaug}; DeepAugment \citep{hendrycks_2021_iccv_manyfaces}; RandConv \citep{xu_2021_iclr_randconv}.}
    \label{fig:04:searching:augmentation:augs}
\end{figure}

In addition to the RandAugment-like augmentations which are composed of simple image processing operations such as Gaussian noise and contrast adjustment, three additional operations are explored that were designed specifically to address the domain gap issue. The effects of these operations are visualized in \cref{fig:04:searching:augmentation:augs}.

\subsubsection{Style Augmentation}

\citet{geirhos_2019_iclr_imagenetbias} has empirically shown in 2019 that many state-of-the-art CNN architectures, such as ResNet \citep{he_2016_cvpr_resnet} and VGG \cite{simonyan_2015_iclr_vgg}, exhibit strong bias toward object's texture when trained on ImageNet \cite{russakovsky_2015_ijcv_imagenet}. This finding is very counter-intuitive for humans who would, for example, classify an object or an animal based on its global or a set of local shapes, not based on its texture. To demonstrate such behavior, the authors have created the Stylized-ImageNet (SIN) dataset by applying the AdaIN Neural Style Transfer (NST) pipeline \citep{huang_2017_iccv_adain} to each image in ImageNet with random styles from Kaggle's \texttt{Painter by Numbers} dataset\footnote{\url{https://www.kaggle.com/c/painter-by-numbers/}}. The result shows that when the same CNN architectures are trained on the SIN dataset, the networks not only exceed the performance of those trained on ImageNet, but they also show human-level robustness to previously unseen image distortions, such as noise, contrast change, and high- or low-pass filtering.

Following the same logic, \citet{jackson_2019_iccvw_styleaug} devised style augmentation using the style transfer pipeline of \citet{ghiasi_2017_bmvc_style}. Specifically, in order to facilitate the style randomization process without resorting to individual ``style images,'' they embedded different style images into the vector space using the intermediate feature output of an Inception CNN \citep{szegedy_2015_cvpr_googlenet}. Then, during training, random style vectors are sampled and passed to the style transfer network to ``stylize'' the input image. This work adopts the style augmentation pipeline, using the same style embedding statistics as a source of randomization due to its ease of sampling different styles from the vector space embedding.

\subsubsection{DeepAugment}

DeepAugment \citep{hendrycks_2021_iccv_manyfaces} is another NN-based data augmentation technique that is done by perturbing the internal representations of a NN. These perturbations are randomly sampled from a set of hand-designed operations which include zeroing, negating, convolving, flipping, etc., of intermediate features and weights of a NN (e.g., an autoencoder) at random layers. For simplicity, this work follows the procedure identical to the one described in \citet{hendrycks_2021_iccv_manyfaces} and uses the CAE architecture \citep{theis_2017_iclr_cae} to produce perturbed images.

\subsubsection{RandConv}

Finally, RandConv \citep{xu_2021_iclr_randconv} applies a convolution operation with a $k \times k$ kernel composed of random weights. The motivation is similar to that of style augmentation -- to distort the local texture of an input image. Following the original description, this work applies a convolution at the beginning with kernel size randomly sampled from $k = \{1, 3, 5, 7\}$, and the perturbed image is blended with the original image with random weights.

\section{Trade-Off Analyses: OOD Robustness and Latency}
\label{sec:05_exp_ood_latency}

This section performs the trade-off analyses between robustness across domain gap and training/inference latency for various configurations of pose estimation architectures (\cref{sec:04_searching:arch}), data augmentation techniques (\cref{sec:04_searching:augment}), and other aspects of NN training such as transfer learning and input resolution. The ultimate goal is to find the configuration that is both robust on SPEED+ HIL domain images and computationally efficient on a representative hardware, capable of operating above nominal GN\&C update frequency (\cref{sec:03_req:efficiency}) while also being batch-agnostic for potential online training (\cref{sec:03_req:batch_agnostic}).

\subsection{Datasets}
\label{sec:05_exp_ood_latency:datasets}

Following SPEC2021 \citep{park_2023_acta_spec2021}, this section performs experiments using the SPEED+ dataset \citep{park_2022_aero_speedplus, park_2021_sdr_speedplus} which consists of 59,960 $\synthetic$ images split in 80:20 ratio into the training and validation sets.
It further includes two HIL image domains acquired from the TRON facility and a mockup model of the Tango spacecraft: 6,740 $\lightbox$ and 2,791 $\sunlamp$ images which are reserved solely for testing. In this section, all NN models are trained on the $\synthetic$ training set and evaluated on the $\synthetic$ validation set, $\lightbox$ and $\sunlamp$ test sets, and 25 labeled flight images acquired in-space during RPO of the PRISMA mission (\prisma). Recall \cref{fig:01:tron_HIL_images} for visualization of a few HIL domain images.

\subsection{Implementations}
\label{sec:05_exp_ood_latency:implementations}

In the original EfficientDet implementation \citep{tan_2020_cvpr_efficientdet}, the input image resolution and the size of the BiFPN and prediction layers differ depending on the EfficientNet scaling parameter $\phi$. In this section, for a fair comparison, the BiFPN layers are fixed to 4 convolution layers with 128 channels. The prediction heads for all pose estimation architectures consist of two convolution layers with 128 channels as well. Note that for most EfficientDet and HRNet implementations, the convolution operations in the prediction heads are depthwise separable convolutions \citep{sandler_2018_cvpr_mobilenetv2} except ViTPose architectures that adopt full transposed 2D convolution operations with $\times$2 upsampling and 128 channels. The $\norm$ and activation layers for BiFPN and prediction heads all follow those used in the respective backbone NN. If the backbone consists of multiple types of $\norm$ layers including BN, the following BiFPN and prediction heads adopt only BN. The 2D keypoint locations are extracted from each heatmap as the location with maximum pixel intensity then refined via Unbiased Data Processing (UDP) \citep{huang_2020_cvpr_udp}. The keypoint coordinates are used to solve for the 6D pose solution via EP$n$P \citep{lepetit_2008_ijcv_epnp}.

Given the ongoing interest in Artificial Intelligence (AI) for space applications, this work assumes that hardware support for edge computing to run NN onboard satellites will increase in the upcoming years (see \cref{sec:03_req:efficiency:gpu}). Therefore, this work adopts an NVIDIA Jetson Nano 4GB which contains a Quad-Core Arm\textsuperscript{\textregistered} Cortex\textsuperscript{\textregistered} A57 CPU and a 128-core NVIDIA Maxwell\texttrademark~architecture GPU \citep{nvidia_jetson_nano}. Nano is the most restrictive version of the commercially available Jetson family, which suits the limited processing power of onboard GPUs as evidenced in the NASA small spacecraft technology report \citep{nasa_soa_smallsat_2023}. Given its deprecated GPU architecture and CUDA language support, all NN training and inference on Jetson Nano are performed with full 32-bit Floating Point (FP32) precision using the C++ API of PyTorch 1.10.0.

\subsubsection{Default Configuration}
\label{sec:05_exp_ood_latency:implementations:default_config}

The default training configuration consists of the heatmap prediction head from the P2 level feature output and all data augmentations including style augmentation, DeepAugment and RandConv. The ground-truth heatmaps are Gaussian blobs with the standard deviation $\sigma = 1$ pixel around the 2D keypoint locations. The NN is trained with the pixel-wise Mean Squared Error (MSE) loss defined for the predicted heatmap $\hat{\bm{\mathcal{H}}}$ and the ground-truth $\bm{\mathcal{H}}$ as
\begin{align} \label{eqn:04:experiments_ood_latency:mse_loss}
    \mathcal{L}_\text{MSE}(\hat{\bm{\mathcal{H}}}, \bm{\mathcal{H}}) = \frac{1}{K} \sum_{i=1}^K \| \hat{\bm{\mathcal{H}}}^{(i)} - \bm{\mathcal{H}}^{(i)} \|_F^2
\end{align}
where $K$ is the number of total heatmaps and $\|\cdot\|_F$ is a Frobenius norm of the heatmap matrix.

All training sessions excluding the experiments on the Jetson Nano are run on an NVIDIA RTX 4090 24GB GPU with \texttt{bfloat16} precision on PyTorch 2.1. All models are trained with the AdamW optimizer \citep{loshchilov_2019_iclr_adamw} for 30 epochs with the batch size 32, weight decay $1.0 \times 10^{-4}$ and the initial learning rate 0.001 which linearly warms up during the first epoch then decays according to the cosine annealing schedule \citep{loshchilov_2017_iclr_sgdr}. Note that the proposed training algorithm is not optimal for all different models considered in this study; nevertheless, adopting a common training method allows for a fair comparison of different metrics.

The default input image resolution after cropping is 224 $\times$ 224, identical to the ImageNet classification scenario. The input RoI is cropped using the ground-truth bounding box around the target following the procedure described in \citet{park_2019_aas_krn}. Specifically, the ground-truth RoI is first corrected to a square-sized region with a size of $\max(h, w)$, where $(h, w)$ are the height and width of the original RoI. This correction is implemented to ensure the aspect ratio of the target remains the same when resizing the cropped region. Then, during training, the new square RoI is enlarged and shifted by random factors in order to make the NN robust to object translation and misaligned RoI detection. During testing, the detected RoI is similarly converted into a square-sized region and enlarged by a fixed factor of 20\% in order to ensure that the cropped region contains the entirety of the spacecraft. 

Finally, all backbone NNs are obtained from the \texttt{timm} library available at HuggingFace \citep{wightman_2019_timm}, and they are all pre-trained on ImageNet-1K (IN-1K) \citep{russakovsky_2015_ijcv_imagenet} alone via supervised training unless noted otherwise. For simplicity, the architecture and backbone composition of a model is noted Architecture (\emph{Backbone}) throughout this section.

\subsubsection{Metrics}
\label{sec:05_exp_ood_latency:implementations:metrics}

The performance metrics consist of the mean translation error ($E_\text{t}$), mean orientation error ($E_\text{q}$) and the mean pose error ($E_\text{pose}$) defined below for individual samples
\begin{subequations}
\begin{align}
    e_\text{t} &= \| \tilde{\bm{t}} - \bm{t} \|_2 \\
    \bar{e}_\text{t} &= e_\text{t} / \|\bm{t}\| \\
    e_\text{q} &= 2 \arccos | < \tilde{\bm{q}}, \bm{q} > | \\
    e_\text{pose} &= e_\text{q} + \bar{e}_\text{t}
\end{align}
\end{subequations}
where $(\bm{t},\bm{q})$ and $(\tilde{\bm{t}}, \tilde{\bm{q}})$ respectively denote the ground-truth and estimated poses comprising 3D translation and 4D quaternion vectors. Also note that for the SPEED+ HIL domains, the rotation and translation errors account for the calibration accuracy of the TRON facility. Specifically, for individual samples, the HIL errors are defined as \citep{park_2023_acta_spec2021}
\begin{align} 
    \bar{e}_\text{t}^*(\tilde{\bm{t}}, \bm{t}) = \begin{cases} 0 & \textrm{if } \bar{e}_\text{t}(\tilde{\bm{t}}, \bm{t}) < \theta_t \\ \bar{e}_\text{t}(\tilde{\bm{t}}, \bm{t}) & \textrm{otherwise} \end{cases}, ~~~
    e_\text{q}^*(\tilde{\bm{q}}, \bm{q}) = \begin{cases} 0 & \textrm{if } e_\text{q}(\tilde{\bm{q}}, \bm{q}) < \theta_q \\ e_\text{q}(\tilde{\bm{q}}, \bm{q}) & \textrm{otherwise} \end{cases}
\end{align}
The rotation threshold ($\theta_q$) and the normalized translation threshold ($\theta_t$) are derived from the calibration accuracy of TRON and are set to $\theta_q = 0.169^\circ$ and $\theta_t = 2.173 \times 10^{-3}$ \citep{park_2021_aas_tron}. Then, the HIL pose error for a sample is given as
\begin{align}
	e_\text{pose}^* = e_\text{q}^*(\tilde{\bm{q}}, \bm{q}) + \bar{e}_\text{t}^*(\tilde{\bm{t}}, \bm{t})
\end{align}

Note that in the overall pose estimation scenario (\cref{sec:035_scenario}), the information on the relative position of the target is mostly included in the RoI detected from either the navigation filter or a separate object detection NN. Since the ground-truth RoIs are used for image pre-processing in this section, mean translation errors are expected to be very small throughout the experiments. Therefore, only the mean orientation errors ($E_\text{q}$ / $E_\text{q}^*$) are reported for the majority of this section for brevity.

\subsection{Ablation Studies}
\label{sec:05_exp_ood_latency:ablation}

\subsubsection{Architecture \& Backbone}
\label{sec:05_exp_ood_latency:ablation:arch}

This section begins with the study on OOD robustness and latency of various pose estimation architectures and NN backbones. First, \cref{tab:05:ablation:arch} tabulates the latency and pose accuracy of different pose estimation architectures with various backbone NNs of different sizes. These backbones cover a wide variety of NN architectures, including conventional CNNs with BN layers (e.g., EfficientNet \citep{tan_2019_icml_efficientnet}, EfficientNetV2 \citep{tan_2021_icml_efficientnetv2}, MobileNetV3 \citep{howard_2019_iccv_mobilenetv3}, HRNet \citep{wang_2021_tpami_hrnet}); NFNet \citep{brock_2021_icml_nfnet} which has no separate $\norm$ layers but performs weight standardization of the convolution weights \citep{qiao_2020_weightstandardization}; Vision Transformers (ViT) \citep{dosovitskiy_2021_iclr_vit, touvron_2021_icml_deit, touvron_2022_eccv_deit3} with self-attention modules \citep{vaswani_2017_nips_transformer}; architectures purely based on Multi-Layer Perceptrons (MLP) (e.g., ResMLP \citep{touvron_2023_tpami_resmlp}); and hybrid architectures which both include convolution operations and self-attention layers for image patches as hinted by the use of both BN/LN layers (e.g., HRFormer \citep{yuan_2021_nips_hrformer}, EfficientFormerV2 \citep{li_2023_iccv_efficientformerv2}, MobileViTv2 \citep{mehta_2023_tmlr_mobilevitv2}.

\begin{table*}[!t]
\caption{Ablation study on different pose estimation architectures and backbones. Latency and peak memory during training are measured on an NVIDIA RTX 4090 24GB GPU with batch size 1 and FP32 precision. Mean{\footnotesize{(std.~dev.)}} across all samples for each domain are reported. The boldface denotes the best performance within each group.}
\label{tab:05:ablation:arch}
\centering
\footnotesize
\tabcolsep=0.1cm 
\begin{tabular}{lccccccccc}
\toprule
\multirow{2}{*}{Backbone} & \multirow{2}{*}{Norm.} & \multirow{2}{*}{\makecell{Num. \\ Param.}} & \multirow{2}{*}{\makecell{Mem. \\ $[$MB$]$}} & \multicolumn{2}{c}{Time [ms]} & \multicolumn{4}{c}{$E_\text{q}$ / $E_\text{q}^*$ [${}^\circ$]} \\
\cmidrule(lr){5-6} \cmidrule(lr){7-10}
& & & & Train & Test & \synthetic & \lightbox & \sunlamp & \prisma \\
\midrule
\tikzcircle{red} MobileNetV3-L ($\times$0.75) \citep{howard_2019_iccv_mobilenetv3} & BN & 2.3M & 149 & 21.3 & 6.1 & \mnstd{1.00}{3.00} & \mnstd{10.48}{29.97} & \mnstd{19.43}{40.25} & \mnstd{3.66}{7.33} \\ 
\tikzcircle{green} HRFormer-T \citep{yuan_2021_nips_hrformer} & BN/LN & 2.6M & 264 & 58.5 & 20.4 & \mnstd{1.22}{4.30} & \mnstd{9.40}{27.30} & \bfseries \mnstd{14.17}{32.64} & \mnstd{17.39}{46.83} \\ 
\tikzcircle{green} HRNet-W18-S \citep{wang_2021_tpami_hrnet} & BN & 2.9M & \bfseries 65 & \bfseries 13.2 & \bfseries 4.8 & \mnstd{1.55}{5.77} & \mnstd{14.68}{33.54} & \mnstd{26.13}{43.51} & \mnstd{22.68}{38.51} \\ 
\tikzcircle{red} MobileNetV3-L ($\times$1.0) \citep{howard_2019_iccv_mobilenetv3} & BN & 3.5M & 168 & 20.4 & 6.1 & \mnstd{0.91}{2.82} & \mnstd{8.19}{25.45} & \mnstd{14.29}{32.96} & \bfseries \mnstd{2.46}{1.76} \\ 
\tikzcircle{red} EfficientFormerV2-S0 \citep{li_2023_iccv_efficientformerv2} & BN/LN & 3.7M & 197 & 25.8 & 7.9 & \mnstd{1.09}{4.07} & \mnstd{10.76}{30.10} & \mnstd{17.08}{36.77} & \mnstd{2.91}{3.90} \\ 
\tikzcircle{red} EfficientNet-B0 \citep{tan_2019_icml_efficientnet} & BN & 4.1M & 212 & 23.1 & 6.6 & \bfseries \mnstd{0.80}{1.88} & \bfseries \mnstd{7.75}{25.41} & \mnstd{14.70}{35.11} & \mnstd{8.61}{30.96} \\ 
\midrule 
\tikzcircle{red} MobileViTv2-100 \citep{mehta_2023_tmlr_mobilevitv2} & BN/LN & 5.0M & 262 & 21.8 & 6.9 & \mnstd{0.81}{2.28} & \mnstd{9.79}{29.06} & \mnstd{17.89}{38.72} & \mnstd{6.55}{16.98} \\ 
\tikzcircle{red} EfficientNetV2-B0 \citep{tan_2021_icml_efficientnetv2} & BN & 6.1M & 208 & 25.9 & 7.3 & \mnstd{0.84}{2.75} & \mnstd{7.25}{24.30} & \mnstd{12.55}{32.43} & \mnstd{2.81}{4.38} \\ 
\tikzcircle{blue} ViT-T/16 \citep{touvron_2022_eccv_deit3} & LN & 6.2M & \bfseries 129 & \bfseries 13.2 & \bfseries 3.8 & \mnstd{1.01}{2.60} & \mnstd{7.88}{23.48} & \mnstd{14.49}{32.25} & \mnstd{16.56}{46.06} \\ 
\tikzcircle{red} EfficientFormerV2-S1 \citep{li_2023_iccv_efficientformerv2} & BN/LN & 6.2M & 252 & 31.5 & 8.7 & \mnstd{1.02}{3.86} & \mnstd{9.50}{27.77} & \mnstd{16.46}{36.21} & \mnstd{17.21}{41.00} \\ 
\tikzcircle{red} EfficientNet-B1 \citep{tan_2019_icml_efficientnet} & BN & 6.6M & 274 & 28.4 & 7.7 & \bfseries \mnstd{0.71}{1.06} & \bfseries \mnstd{6.07}{21.58} & \bfseries \mnstd{9.34}{26.40} & \bfseries \mnstd{2.50}{2.79} \\ 
\midrule 
\tikzcircle{red} EfficientNetV2-B1 \citep{tan_2021_icml_efficientnetv2} & BN & 7.1M & \bfseries 236 & 29.2 & 7.7 & \mnstd{0.78}{2.93} & \mnstd{5.87}{21.20} & \mnstd{9.44}{27.60} & \mnstd{3.53}{6.99} \\ 
\tikzcircle{red} MobileViTv2-125 \citep{mehta_2023_tmlr_mobilevitv2} & BN/LN & 7.5M & 324 & \bfseries 21.4 & \bfseries 6.8 & \mnstd{0.75}{2.82} & \mnstd{7.92}{25.56} & \mnstd{10.87}{29.03} & \mnstd{3.48}{6.66} \\ 
\tikzcircle{red} EfficientNet-B2 \citep{tan_2019_icml_efficientnet} & BN & 7.7M & 292 & 27.6 & 7.9 & \bfseries \mnstd{0.70}{2.02} & \bfseries \mnstd{5.21}{19.44} & \bfseries \mnstd{8.05}{23.99} & \bfseries \mnstd{1.98}{1.52} \\ 
\midrule 
\tikzcircle{green} HRFormer-S \citep{yuan_2021_nips_hrformer} & BN/LN & 7.9M & 500 & 69.2 & 23.7 & \mnstd{0.94}{3.07} & \mnstd{6.77}{22.17} & \mnstd{10.60}{27.68} & \mnstd{37.88}{62.67} \\ 
\tikzcircle{red} EfficientNetV2-B2 \citep{tan_2021_icml_efficientnetv2} & BN & 8.9M & \bfseries 263 & 30.3 & 8.1 & \mnstd{0.73}{2.18} & \mnstd{5.63}{20.59} & \mnstd{8.48}{25.36} & \mnstd{1.92}{1.39} \\ 
\tikzcircle{red} MobileViTv2-150 \citep{mehta_2023_tmlr_mobilevitv2} & BN/LN & 10.5M & 393 & \bfseries 21.9 & \bfseries 6.8 & \mnstd{0.71}{2.27} & \mnstd{7.85}{25.46} & \mnstd{10.19}{27.72} & \bfseries \mnstd{1.88}{1.09} \\ 
\tikzcircle{red} EfficientNet-B3 \citep{tan_2019_icml_efficientnet} & BN & 10.6M & 369 & 27.9 & 8.5 & \bfseries \mnstd{0.66}{2.16} & \bfseries \mnstd{4.67}{18.19} & \bfseries \mnstd{7.08}{21.24} & \mnstd{2.07}{1.78} \\ 
\midrule 
\tikzcircle{green} HRNet-W18 \citep{wang_2021_tpami_hrnet} & BN & 11.0M & \bfseries 238 & 42.4 & 13.7 & \mnstd{0.88}{2.79} & \mnstd{6.26}{20.92} & \mnstd{10.52}{28.54} & \mnstd{16.10}{47.52} \\ 
\tikzcircle{red} EfficientFormerV2-S2 \citep{li_2023_iccv_efficientformerv2} & BN/LN & 12.6M & 374 & 34.3 & 11.4 & \mnstd{0.93}{4.13} & \mnstd{6.94}{23.33} & \mnstd{11.94}{31.53} & \mnstd{2.13}{1.63} \\ 
\tikzcircle{red} EfficientNetV2-B3 \citep{tan_2021_icml_efficientnetv2} & BN & 12.9M & 335 & \bfseries 29.1 & \bfseries 8.9 & \mnstd{0.69}{2.18} & \bfseries \mnstd{5.11}{19.08} & \bfseries \mnstd{7.41}{23.00} & \mnstd{3.48}{7.02} \\ 
\tikzcircle{red} EfficientNet-B4 \citep{tan_2019_icml_efficientnet} & BN & 17.2M & 499 & 31.4 & 9.9 & \bfseries \mnstd{0.69}{1.47} & \mnstd{5.67}{20.27} & \mnstd{7.79}{23.76} & \bfseries \mnstd{1.94}{1.12} \\ 
\midrule 
\tikzcircle{red} EfficientNetV2-S \citep{tan_2021_icml_efficientnetv2} & BN & 20.3M & 468 & 32.9 & 10.3 & \mnstd{0.62}{1.72} & \mnstd{4.24}{16.63} & \mnstd{6.10}{18.91} & \mnstd{4.55}{13.94} \\ 
\tikzcircle{red} NFNet-L0 \citep{brock_2021_icml_nfnet} & - & 22.3M & 497 & 24.3 & 8.3 & \mnstd{0.64}{2.06} & \mnstd{5.59}{20.19} & \mnstd{7.91}{23.41} & \mnstd{6.25}{16.25} \\ 
\tikzcircle{blue} ViT-S/16 \citep{touvron_2022_eccv_deit3} & LN & 22.7M & \bfseries 377 & \bfseries 11.8 & \bfseries 3.3 & \mnstd{0.75}{1.65} & \mnstd{5.12}{17.75} & \mnstd{9.29}{24.94} & \mnstd{2.32}{1.36} \\ 
\tikzcircle{green} HRNet-W30 \citep{wang_2021_tpami_hrnet} & BN & 27.4M & 493 & 44.8 & 13.6 & \mnstd{0.73}{2.09} & \mnstd{5.16}{19.22} & \mnstd{8.19}{24.52} & \mnstd{6.78}{23.84} \\ 
\tikzcircle{red} EfficientNet-B5 \citep{tan_2019_icml_efficientnet} & BN & 27.8M & 707 & 39.3 & 11.1 & \bfseries \mnstd{0.58}{1.37} & \bfseries \mnstd{3.79}{15.59} & \bfseries \mnstd{5.26}{16.63} & \bfseries \mnstd{2.12}{1.36} \\ 
\midrule 
\tikzcircle{blue} ResMLP-S24 \citep{touvron_2023_tpami_resmlp} & Affine & 30.7M & \bfseries 604 & 16.4 & 4.2 & \mnstd{1.08}{3.21} & \mnstd{7.33}{21.62} & \mnstd{18.65}{37.93} & \mnstd{2.18}{1.17} \\ 
\tikzcircle{red} NFNet-L1 \citep{brock_2021_icml_nfnet} & - & 37.6M & 799 & 33.5 & 12.7 & \mnstd{0.54}{1.82} & \mnstd{4.68}{18.76} & \mnstd{6.53}{21.80} & \mnstd{3.64}{8.76} \\
\tikzcircle{blue} ViT-M/16 \citep{touvron_2022_eccv_deit3} & LN & 39.6M & 647 & \bfseries 11.5 & \bfseries 3.4 & \mnstd{0.70}{2.11} & \mnstd{4.32}{15.77} & \mnstd{8.11}{22.57} & \mnstd{2.18}{1.07} \\ 
\tikzcircle{red} EfficientNet-B6 \citep{tan_2019_icml_efficientnet} & BN & 39.9M & 944 & 39.9 & 11.9 & \bfseries \mnstd{0.53}{1.00} & \bfseries \mnstd{3.65}{15.14} & \bfseries \mnstd{5.19}{18.03} & \bfseries \mnstd{1.99}{1.06} \\ 
\bottomrule
\multicolumn{10}{l}{\tikzcircle{red} EfficientDet ~~ \tikzcircle{green} HRNet ~~ \tikzcircle{blue} ViTPose}
\end{tabular}
\end{table*}

In order to facilitate readability, \cref{tab:05:ablation:arch} sorts different architectures according to their total number of parameters in ascending order and grouped into similar sizes. Overall, larger NNs tend to result in better OOD robustness as measured by the performance on the SPEED+ $\lightbox$ and $\sunlamp$ domains. Then, by investigating the models within each size group, it becomes immediately clear that the EfficientDet (\emph{EfficientNet}) and EfficientDet (\emph{EfficientNetV2}) consistently outperform all other models across different size groups in terms of robustness. They are closely followed by EfficientDet (\emph{NFNet}) and ViTPose (\emph{ViT}). 

\begin{figure}[!t]
    \centering
    \begin{subfigure}[b]{0.4\linewidth}
        \centering
        \includegraphics[width=1.0\textwidth]{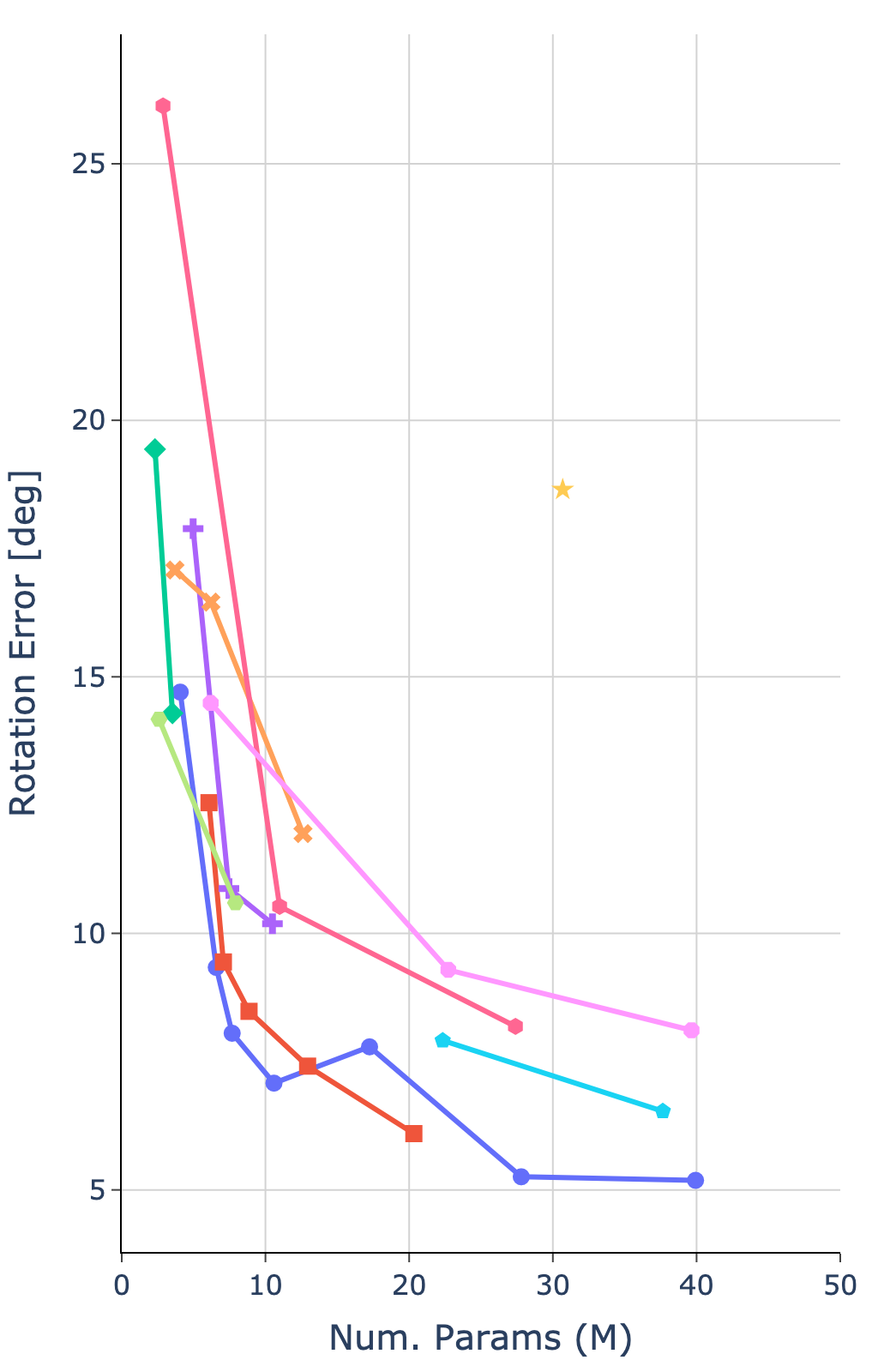}
    \end{subfigure}
    \begin{subfigure}[b]{0.4\linewidth}
        \centering
        \includegraphics[width=1.0\textwidth]{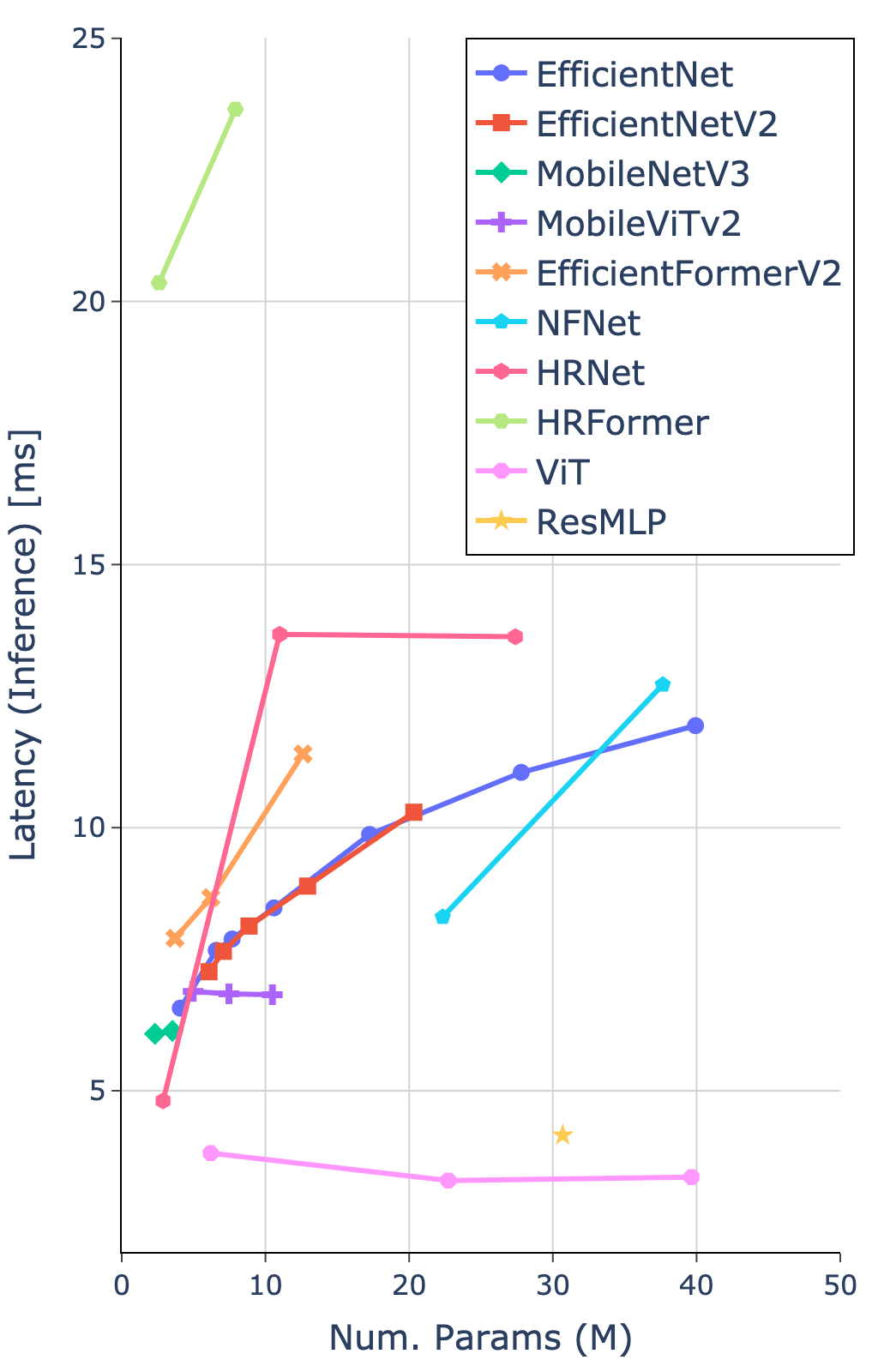}
    \end{subfigure}
    \caption{Visualization of performance of different backbones tabulated in \cref{tab:05:ablation:arch}. Rotation error ($E_\text{q}^*$) is for the $\sunlamp$ domain.}
    \label{fig:05:ablation:arch:param_vs_sunlamp_time}
\end{figure}

\Cref{fig:05:ablation:arch:param_vs_sunlamp_time} condenses the information in \cref{tab:05:ablation:arch} by plotting the mean orientation error on the $\sunlamp$ domain and the inference latency against the number of parameters. Different families of backbones are plotted with different colors and marker shapes. First, the $E_\text{q}^*$ plot in \cref{fig:05:ablation:arch:param_vs_sunlamp_time} (\emph{left}) shows that EfficientDet (\emph{EfficientNet}) and EfficientDet (\emph{EfficientNetV2}) consistently outperform all other models in terms of OOD robustness. While EfficientDet (\emph{NFNet}) models are closely behind them, there is no smaller variant of NFNet pre-trained on IN-1K that is publicly available. On the other hand, the latency plot in \cref{fig:05:ablation:arch:param_vs_sunlamp_time} (\emph{right}) shows that ViTPose (\emph{ViT}) and ViTPose (\emph{ResMLP}) are clear winners. Both architectures consist solely of MLP layers, enabling a faster throughput compared to those with convolution operations. Unfortunately, ResMLP cannot perform well across domain gaps, which is no surprise since it is a pure MLP-based NN with no inductive bias of convolution operations or global spatial knowledge via self-attention layers. However, ViTPose (\emph{ViT}) models closely follow EfficientDet (\emph{EfficientNet}) in terms of OOD robustness.

Based on the OOD robustness and latency trade-off in \cref{fig:05:ablation:arch:param_vs_sunlamp_time},  only EfficientDet (\emph{EfficientNet}), EfficientDet (\emph{EfficientNetV2}) and ViTPose (\emph{ViT}) are considered for further analyses.

\subsubsection{Latency on NVIDIA Jetson Nano}
\label{sec:05_exp_ood_latency:ablation:latency_jetson}

\begin{table*}[!t]
\caption{Peak training memory and latency of select models from \cref{tab:05:ablation:arch} are measured on an onboard GPU of NVIDIA Jetson Nano 4GB with PyTorch C++ API for batch size 1 and FP32 precision. The boldface denotes the best performance within each group.}
\label{tab:05:ablation:latency}
\centering
\footnotesize
\tabcolsep=0.1cm 
\begin{tabular}{lccccccc}
\toprule
\multirow{2}{*}{Backbone} & \multirow{2}{*}{Norm.} & \multirow{2}{*}{\makecell{Num. \\ Param.}} & \multirow{2}{*}{\makecell{Mem. \\ $[$MB$]$}} & \multicolumn{2}{c}{Latency [ms]} & \multicolumn{2}{c}{$E_\text{q}^*$ [${}^\circ$]} \\
\cmidrule(lr){5-6} \cmidrule(lr){7-8}
& & & & Train & Test & \lightbox & \sunlamp \\
\midrule
\tikzcircle{red} EfficientNet-B0 & BN & 4.1M & \bfseries 212 & \bfseries 446 & \bfseries 57 & \mnstd{7.75}{25.41} & \mnstd{14.70}{35.11} \\ 
\tikzcircle{red} EfficientNet-B1 & BN & 6.6M & 274 & 595 & 72 & \mnstd{6.07}{21.58} & \mnstd{9.34}{26.40} \\ 
\tikzcircle{red} EfficientNet-B2 & BN & 7.7M & 292 & 622 & 71 & \mnstd{5.21}{19.44} & \mnstd{8.05}{23.99} \\ 
\tikzcircle{red} EfficientNet-B3 & BN & 10.6M & 369 & 758 & 75 & \mnstd{4.67}{18.19} & \mnstd{7.08}{21.24} \\ 
\tikzcircle{red} EfficientNet-B4 & BN & 17.2M & 499 & 985 & 87 & \mnstd{5.67}{20.27} & \mnstd{7.79}{23.76} \\ 
\tikzcircle{red} EfficientNet-B5 & BN & 27.8M & 707 & 1317 & 98 & \mnstd{3.79}{15.59} & \mnstd{5.26}{16.63} \\ 
\tikzcircle{red} EfficientNet-B6 & BN & 39.9M & 944 & 1660 & 110 & \bfseries \mnstd{3.65}{15.14} & \bfseries \mnstd{5.19}{18.03} \\ 
\midrule
\tikzcircle{red} EfficientNetV2-B0 & BN & 6.1M & \bfseries 208 & \bfseries 473 & \bfseries 65 & \mnstd{7.25}{24.30} & \mnstd{12.55}{32.43} \\ 
\tikzcircle{red} EfficientNetV2-B1 & BN & 7.1M & 236 & 553 & 72 & \mnstd{5.87}{21.20} & \mnstd{9.44}{27.60} \\ 
\tikzcircle{red} EfficientNetV2-B2 & BN & 8.9M & 263 & 598 & 74 & \mnstd{5.63}{20.59} & \mnstd{8.48}{25.36} \\ 
\tikzcircle{red} EfficientNetV2-B3 & BN & 12.9M & 335 & 746 & 82 & \mnstd{5.11}{19.08} & \mnstd{7.41}{23.00} \\ 
\tikzcircle{red} EfficientNetV2-S & BN & 20.3M & 468 & 996 & 96 & \bfseries \mnstd{4.24}{16.63} & \bfseries \mnstd{6.10}{18.91} \\ 
\midrule
\tikzcircle{blue} ViT-T/16 & LN & 6.2M & \bfseries 129 & \bfseries 221 & 25 & \mnstd{7.88}{23.48} & \mnstd{14.49}{32.25}\\
\tikzcircle{blue} ViT-S/16 & LN & 22.7M & 377 & 349 & \bfseries 22 & \mnstd{5.12}{17.75} & \mnstd{9.29}{24.94} \\ 
\tikzcircle{blue} ViT-M/16 & LN & 39.6M & 647 & 546 & 23 & \bfseries \mnstd{4.32}{15.77} & \bfseries \mnstd{8.11}{22.57} \\ 
\bottomrule
\multicolumn{8}{l}{\tikzcircle{red} EfficientDet ~~ \tikzcircle{blue} ViTPose}
\end{tabular}
\end{table*}

The models chosen from the previous section are implemented on an NVIDIA Jetson Nano 4GB and tested for training and inference latency on its GPU. Specifically, all training and inference are performed with full FP32 precision for single images. The FP32 training and inference are due to the lack of support for FP16 or INT8 quantization on the Nano's GPU, whereas a single image training is due to the fact that it is undesirable for a satellite to wait and collect a batch of images of the target spacecraft during in-space RPOD, both logistically and computationally.

The latency measured on the Jetson Nano reported in \cref{tab:05:ablation:latency} reaffirms the observation made in \cref{tab:05:ablation:arch} and \cref{fig:05:ablation:arch:param_vs_sunlamp_time}. For the same number of parameters, EfficientDet models dominate on the OOD robustness front compared to the ViTPose models. However, ViTPose (\emph{ViT}) models are consistently faster for inference regardless of the size. In fact, the largest ViT backbone considered in this study---ViT-M/16 with nearly 40M parameters---runs nearly 2.5 times faster than the smallest variant of the EfficientNet backbone that is 1/10 in size, and its training latency is on par with that of EfficientNet-B1 which only has 6.6M parameters. It also requires at most about 650 MB of GPU memory for each training session with a single image input. Considering the 4GB of GPU memory available on the Jetson Nano and assuming that most of the GN\&C-related operations run on CPUs, 650 MB or less is completely within the range of operation excluding the possibility of running other image-processing algorithms on the GPU.

Recall from the earlier discussion that the training is expected to run only occasionally whenever there has been a substantial change in the scenery and the target's pose (\cref{sec:03_req:efficiency}). Therefore, even the EfficientNet-B6 backbone could perform onboard online training asynchronously on a GPU given that there is enough power generated to support such a computationally heavy operation. In terms of the inference speed, all models considered in \cref{tab:05:ablation:latency} can effectively run in real-time for 0.5 Hz GN\&C updates of the CPOD reference mission \citep{roscoe_2018_acta_cpod} assuming that the pre- and post-processing of the NN inputs and outputs do not become the computational bottleneck.

\subsubsection{Replacing BN Layers}
\label{sec:05_exp_ood_latency:ablation:replace_bn}

Judging from the comparison on the NVIDIA Jetson Nano reported in \cref{tab:05:ablation:latency}, it appears that EfficientDet models are more OOD robust for the same number of parameters, but the larger ViTPose models are still more computationally efficient than small EfficientDet models, overcoming its comparative weakness in OOD robustness with increased size. There is one more reason that ViTPose is favored over EfficientDet, and that is due to the requirement that the onboard NNs must be batch-agnostic for online training (\cref{sec:03_req:batch_agnostic}). As evidenced in \cref{tab:05:ablation:arch}, most CNN backbones designed for ImageNet classification adopt BN layers, and even hybrid architectures always pair BN layers to each convolution operation. Then, is it possible to replace those BN layers with other batch-agnostic layers (e.g., LN, GN) when the overwhelming majority of publicly available CNN backbones pre-trained on ImageNet come with BN layers?

\begin{table*}[!t]
\caption{EfficientDet architecture with the EfficientNet-B0 backbone evaluated with different $\norm$ layers. The architecture's and the backbone's $\norm$ layers during IN-1K pretraining are both noted. Mean{\footnotesize{(std.~dev.)}} of average performances over 3 training sessions with different random seeds are reported. Latency is measured on an NVIDIA Jetson Nano 4G. The boldface denotes the best performance within each group.}
\label{tab:05:ablation:norm_layer}
\centering
\footnotesize
\begin{tabular}{cccccccccc}
\toprule
\multirow{2}{*}{\makecell{$\norm$ \\ (Pre-train)}} & \multirow{2}{*}{\makecell{$\norm$ \\ (Current)}} & \multirow{2}{*}{\makecell{Training \\ Epochs}} & \multicolumn{2}{c}{Time {[}ms{]}} & \multicolumn{4}{c}{$E_\text{q}$ / $E_\text{q}^*$ [${}^\circ$]} \\
\cmidrule(lr){4-5} \cmidrule(lr){6-9}
& & & Train & Test & \synthetic & \lightbox & \sunlamp & \prisma \\
\midrule 
BN & BN & 30 & \bfseries 446 & \bfseries 57  & \bfseries \mnstd{0.81}{0.00} & \mnstd{7.34}{0.27} & \mnstd{13.90}{0.37} & \mnstd{7.54}{3.96} \\
\midrule
BN & GN & 30 & 471 & 68 & \mnstd{1.10}{0.01} & \mnstd{12.44}{0.39} & \mnstd{23.18}{2.01} & \mnstd{12.24}{1.32} \\ 
BN & GN & 60 & 471 & 68 & \mnstd{0.88}{0.02} & \mnstd{9.76}{0.34} & \mnstd{18.16}{1.28} & \mnstd{11.47}{4.33} \\ 
GN & GN & 30 & 471 & 68 & \mnstd{0.84}{0.00} & \bfseries \mnstd{6.67}{0.49} & \bfseries \mnstd{12.20}{0.78} & \bfseries \mnstd{3.99}{2.50} \\ 
\bottomrule
\end{tabular}
\end{table*}

\Cref{tab:05:ablation:norm_layer} reports the orientation error of the EfficientDet (\emph{EfficientNet-B0}) model (4.1M parameters) with different $\norm$ layers during the IN-1K pre-training and the main training on SPEED+ $\synthetic$ images. Note that it reports the mean and standard deviations of \emph{average} rotation errors across three separate training sessions with different random seeds. This is in order to demonstrate that the reported performances are not ``lucky'' results due to some favorable random seeds, but rather all training sessions lead to consistent minima of the training objective. The baseline is using BN layers for both sessions, which reports $E_\text{q} = 6.63^\circ$ on $\lightbox$ and $E_\text{q} = 10.77^\circ$ on $\sunlamp$ across 3 training sessions with different random seeds. When the BN layers of a pre-trained backbone is replaced with GN layers with group size 8, fully inheriting the trained affine parameters of the BN layers, the mean orientation error degrades by more than a factor of two. Training the network longer for 60 epochs instead of 30 does improve the overall performance, but it still does not reach the baseline level for all image domains. This was the conclusion of SPNv2 which also reported degradation in performance after replacing the BN with GN layers \citep{park_2023_asr_spnv2}.

However, when the backbone is pre-trained on IN-1K with GN layers from the very beginning, the OOD performance shows consistent improvement on not only $\lightbox$ and $\sunlamp$ but also $\prisma$. The result indicates that it is possible to replace the BN layers in existing CNN architectures with GN or other batch-agnostic layers, but in order to preserve the OOD robustness, the backbone must be pre-trained on IN-1K or other large-scale datasets with those batch-agnostic layers from the very beginning. The problem is that ImageNet training is extremely expensive, both computationally and in terms of training time. For this study, EfficientNet-B0 was trained on 1M images of IN-1K loosely following the training Recipe B of \citet[Appendix F]{wightman_2021_nips_resnet_recipe} for 350 epochs. Training this small NN on two NVIDIA RTX 4090 24GB GPUs took well over 100 hours. To train larger models would require even more computing power. If such heavy offline training can be afforded, one can use the EfficientNet-B0 backbone with GN layers that show OOD robustness comparable to the ViT-M/16 backbone with 40M parameters at the expense of increased inference time but with much less power consumption. However, if ImageNet pre-training cannot be afforded, it would be preferred to adopt ViT which is available for various ImageNet pre-trained models without BN layers.

\subsubsection{Transfer Learning}
\label{sec:05_exp_ood_latency:ablation:transfer_learning}

\begin{table}[!t]
\caption{Performance of the ViTPose pose estimation architecture with three backbones (ViT-T/16 \& ViT-S/16 \& ViT-M/16) with different pretraining datasets (None \& ImageNet-1K (IN-1K) \& ImageNet-22K (IN-22K)) and methods (DeiT \citep{touvron_2021_icml_deit} \& DeiT III \citep{touvron_2022_eccv_deit3}). Mean{\footnotesize{(std.~dev.)}} of average performances over 3 training sessions with different random seeds are reported. The boldface denotes the best performance within each group.}
\label{tab:05:ablation:transfer}
\centering
\footnotesize
\tabcolsep=0.15cm 
\begin{tabular}{lccccccc}
\toprule
\multirow{2}{*}{\makecell[l]{Backbone \\ (Pre-Train Method)}} & \multirow{2}{*}{\makecell{Pre-Train \\ Dataset}} & \multirow{2}{*}{\makecell{Num. \\ Param.}} & \multirow{2}{*}{\makecell{IN-1K \\ top-1 (\%)}} & \multicolumn{4}{c}{$E_\text{q}$ / $E_\text{q}^*$ [${}^\circ$]} \\
\cmidrule(lr){5-8}
& & & & \synthetic & \lightbox & \sunlamp & \prisma \\
\midrule
\tikzcircle{blue} ViT-T/16 & - & 6.2M & - & \mnstd{16.12}{1.14} & \mnstd{59.72}{0.82} & \mnstd{83.34}{0.71} & \mnstd{66.46}{7.75} \\ 
\tikzcircle{blue} ViT-T/16 (DeiT) & IN-1K & 6.2M & 72.17 & \mnstd{1.13}{0.02} & \mnstd{8.92}{0.38} & \mnstd{18.90}{0.53} & \mnstd{17.67}{3.21} \\
\tikzcircle{blue} ViT-T/16 (DeiT III)\textsuperscript{\textdagger} & IN-1K & 6.2M & 75.12 & \bfseries \mnstd{1.02}{0.01} & \bfseries \mnstd{8.27}{0.09} & \bfseries \mnstd{14.23}{0.47} & \bfseries \mnstd{10.51}{4.80} \\ 
\midrule
\tikzcircle{red} EfficientNetV2-B0 & IN-1K & 6.2M & 78.37 & 0.84 & 7.25 & 12.55 & 2.81 \\ 
\midrule
\tikzcircle{blue} ViT-S/16 & - & 22.7M & - & \mnstd{6.40}{1.35} & \mnstd{41.63}{3.86} & \mnstd{65.74}{2.57} & \mnstd{39.39}{7.96} \\ 
\tikzcircle{blue} ViT-S/16 (DeiT) & IN-1K & 22.7M & 79.86 & \mnstd{0.87}{0.02} & \mnstd{6.11}{0.15} & \mnstd{13.20}{0.89} & \mnstd{8.79}{0.36} \\ 
\tikzcircle{blue} ViT-S/16 (DeiT III) & IN-1K & 22.7M & 81.36 & \mnstd{0.75}{0.00} & \mnstd{4.98}{0.14} & \mnstd{8.81}{0.29} & \mnstd{4.26}{3.08} \\ 
\tikzcircle{blue} ViT-S/16 (DeiT III) & IN-22K & 22.7M & 83.07 & \bfseries \mnstd{0.74}{0.02} & \bfseries \mnstd{4.84}{0.01} & \bfseries \mnstd{8.29}{0.16} & \bfseries \mnstd{2.24}{0.08} \\ 
\midrule
\tikzcircle{red} EfficientNetV2-S & IN-1K & 20.3M & 83.91 & 0.62 & 4.24 & 6.10 & 4.55 \\ 
\midrule
\tikzcircle{blue} ViT-M/16 & - & 39.6M & - & \mnstd{36.19}{1.85} & \mnstd{79.61}{1.13} & \mnstd{95.43}{0.98} & \mnstd{82.80}{9.73} \\ 
\tikzcircle{blue} ViT-M/16 (DeiT III) & IN-1K & 39.6M & 83.10 & \mnstd{0.71}{0.01} & \mnstd{4.46}{0.20} & \mnstd{8.36}{0.15} & \bfseries \mnstd{2.19}{0.06} \\ 
\tikzcircle{blue} ViT-M/16 (DeiT III) & IN-22K & 39.6M & 84.56 & \bfseries \mnstd{0.67}{0.00} & \bfseries \mnstd{4.20}{0.11} & \bfseries \mnstd{7.73}{0.11} & \mnstd{4.18}{2.01} \\ 
\midrule
\tikzcircle{red} EfficientNet-B6 & IN-1K & 39.9M & 84.12 & 0.53 & 3.65 & 5.19 & 1.99 \\ 
\bottomrule
\multicolumn{8}{l}{\tikzcircle{red} EfficientDet ~~ \tikzcircle{blue} ViTPose ~~ \textsuperscript{\textdagger} Trained by the authors} \\
\end{tabular}
\end{table}

The results so far indicate that ViTPose models are superior in terms of computational efficiency and are batch-agnostic by nature, satisfying both Requirements \#1 and \#3 of \cref{sec:03_req}. However, they still fall short in OOD robustness when compared to EfficientDet models of comparable sizes. Therefore, it would be favorable to improve the OOD robustness of ViTPose models without substantial sacrifice of computational advantages. One potential remedy is to improve the transfer learning from the ImageNet pre-training. The ViT backbones of the ViTPose architecture so far are all pre-trained on IN-1K according to the DeiT III recipe \citep{touvron_2022_eccv_deit3}. This is an improved training method compared to DeiT \citep{touvron_2021_icml_deit} which is, in turn, an improved training method compared to the vanilla ViT \citep{dosovitskiy_2021_iclr_vit}. In this section, the effect of ImageNet pre-training for ViT backbones is analyzed in detail by comparing the two different pre-training methods (DeiT vs.~DeiT III) and datasets in \cref{tab:05:ablation:transfer}. First, using random initialization without any pre-training on a large-scale dataset does not train well on SPEED+. For ViT-T/16 and ViT-S/16, using DeiT III leads to improved performance over DeiT when using only IN-1K for supervised pre-training. When the backbone is trained instead on a much larger ImageNet-22K (IN-22K) \citep{russakovsky_2015_ijcv_imagenet} which contains nearly 14M images and 22K class labels then fine-tuned on IN-1K, there is a marginal improvement in OOD robustness for both ViT-S/16 and ViT-M/16 models. The results overall indicate that better generalization of the backbone NN on a large-scale image classification dataset leads to a noticeable improvement in the downstream pose estimation task across the sim2real gap, though it seems the size of the pre-training dataset has little effect on the OOD robustness beyond the scale of IN-1K.

\subsubsection{Input Resolution}
\label{sec:05_exp_ood_latency:ablation:input_resolution}

\begin{table*}[!t]
\caption{Performance of ViTPose (\emph{ViT-S/16}) pre-trained via DeiT III on IN-22K. The input image resolution and the standard deviation ($\sigma$) of the ground-truth heatmaps are varied. Mean{\footnotesize{(std.~dev.)}} of average performances over 3 training sessions with different random seeds are reported. Latency is measured on an NVIDIA Jetson Nano 4GB. The boldface denotes the best performance.}
\label{tab:05:ablation:in_resolution}
\centering
\footnotesize
\tabcolsep=0.15cm 
\begin{tabular}{lcccccccc}
\toprule
\multirow{2}{*}{\makecell[l]{Input Res.}} & \multirow{2}{*}{$\sigma$ [pix]} & \multirow{2}{*}{\makecell{Mem. \\ $[$MB$]$}} & \multicolumn{2}{c}{Latency [ms]} & \multicolumn{4}{c}{$E_\text{q}$ / $E_\text{q}^*$ [${}^\circ$]} \\
\cmidrule(lr){4-5} \cmidrule(lr){6-9}
& & & Train & Test & \synthetic & \lightbox & \sunlamp & \prisma \\
\midrule
224 $\times$ 224 & 1 & \bfseries 377 & \bfseries 349 & \bfseries 22 & \mnstd{0.74}{0.02} & \mnstd{4.84}{0.01} & \mnstd{8.29}{0.16} & \mnstd{2.24}{0.08} \\ 
288 $\times$ 288 & 1 & 401 & 566 & 23 & \mnstd{0.58}{0.01} & \mnstd{3.88}{0.12} & \mnstd{6.70}{0.22} & \mnstd{5.79}{2.82} \\ 
384 $\times$ 384 & 1 & 498 & 1046 & 27 & \mnstd{0.47}{0.01} & \mnstd{3.30}{0.12} & \mnstd{5.61}{0.40} & \mnstd{2.22}{0.17} \\ 
384 $\times$ 384 & 2 & 498 & 1046 & 27 & \mnstd{0.51}{0.00} & \mnstd{2.99}{0.11} & \mnstd{5.11}{0.08} & \mnstd{1.86}{0.01} \\ 
448 $\times$ 448 & 2 & 585 & 1500 & 34 & \bfseries \mnstd{0.47}{0.01} & \bfseries \mnstd{3.16}{0.13} & \bfseries \mnstd{4.53}{0.16} & \bfseries \mnstd{1.78}{0.11} \\ 
\midrule
\multicolumn{9}{l}{\underline{EfficientDet (\emph{EfficientNet-B6})}} \\
224 $\times$ 224 & 1 & 944 & 1660 & 110 & 0.53 & 3.65 & 5.19 & 1.99 \\ 
\bottomrule
\end{tabular}
\end{table*}

Another potential remedy to improve the OOD robustness of ViTPose models is to increase the input image resolution. The effect of input image resolution of ViTPose on the OOD robustness and latency on the Jetson Nano is reported in \cref{tab:05:ablation:in_resolution}. Unsurprisingly, increasing the resolution of the input image cropped around the target leads to improved performance across the board but at the cost of increased peak training memory and latency. Specifically, doubling the input resolution from 224 $\times$ 224 to 448 $\times$ 448 nearly quadruples the training time for each input. This is expected, since doubling the input resolution but maintaining the same patch size (16 $\times$ 16) leads to double the length of patch tokens fed into the transformer blocks, and the memory footprint of the scaled dot-product operation of self-attention modules grows quadratically to the token length. Even though the training latency is tantamount to that of EfficientDet (\emph{EfficientNet-B6}) operating on 224 $\times$ 224 inputs with nearly 40M parameters, the inference latency is still about a third and the training memory nearly half of EfficientDet (\emph{EfficientNet-B6}) while performing better on all image domains. The result suggests that it could be beneficial to use ViT-based models with higher input resolutions as long as the training time does not become the bottleneck.

\subsubsection{Data Augmentation}
\label{sec:05_exp_ood_latency:ablation:data_augmentation}

\begin{table}[!t]
\caption{Performance of the ViTPose-S/16 different different data augmentation configurations. Mean{\footnotesize{(std.~dev.)}} of average performances over 3 training sessions with different random seeds are reported. The boldface denotes the best performance.}
\label{tab:05:ablation:augment}
\centering
\footnotesize
\tabcolsep=0.1cm 
\begin{tabular}{ccccccc}
\toprule
\multirow{2}{*}{SA} & \multirow{2}{*}{DA} & \multirow{2}{*}{RC} & \multicolumn{4}{c}{$E_\text{q}$ / $E_\text{q}^*$ [${}^\circ$]} \\
\cmidrule(lr){4-7}
& & & \synthetic & \lightbox & \sunlamp & \prisma \\
\midrule
- & - & - & \bfseries \mnstd{0.64}{0.01} & \mnstd{6.49}{0.22} & \mnstd{13.18}{0.62} & \mnstd{36.45}{5.28} \\ 
\checkmark & - & - & \mnstd{0.69}{0.01} & \mnstd{6.16}{0.08} & \mnstd{9.80}{0.38} & \mnstd{2.40}{0.21} \\ 
\checkmark & \checkmark & - & \mnstd{0.71}{0.01} & \mnstd{5.66}{0.35} & \mnstd{10.30}{0.75} & \bfseries \mnstd{2.20}{0.16} \\ 
- & \checkmark & \checkmark & \mnstd{0.73}{0.01} & \mnstd{4.84}{0.18} & \mnstd{9.22}{0.08} & \mnstd{3.58}{0.88} \\ 
\checkmark & \checkmark & \checkmark & \mnstd{0.72}{0.01} & \bfseries \mnstd{4.68}{0.27} & \bfseries \mnstd{8.50}{0.15} & \mnstd{2.22}{0.09} \\ 
\bottomrule
\end{tabular}
\end{table}

The experiments so far used all three data augmentation techniques---Style Augmentation, DeepAugment and RandConv. In order to verify the individual contribution of these augmentations, \cref{tab:05:ablation:augment} reports the mean orientation errors for different data augmentation configurations. The default baseline only consists of 5 random augmentations from the Albumentations library implemented in the style of RandAugment \citep{cubuk_2020_nips_randaugment} (\cref{sec:04_searching:augment}).

Note that adding just the style augmentation to the training significantly improves the performance on $\sunlamp$, and incrementally adding DeepAugment and RandConv leads to consistent performance improvement on $\lightbox$. While it does not bear much statistical significance, an interesting observation is that the performances on 25 flight images of $\prisma$ dramatically improve and stabilize (as observed from the standard deviation) once the style augmentation and other techniques are implemented during the training.

\subsection{Summary}
\label{sec:05_exp_ood_latency:final_results}

The previous section investigates different aspects of designing and training a robust pose estimation NN. Under the same training conditions with identical input image resolutions, EfficientDet (\emph{EfficientNet}) consistently outperforms all other models in terms of the OOD robustness as measured on the SPEED+ HIL domains. On the other hand, ViTPose (\emph{ViT}) dominates the training and inference latency for different model sizes but with inferior OOD robustness. However, various ablation studies reveal that the ViTPose models can be made as robust as EfficientDet with improved pre-training of the ViT backbone, extensive data augmentation and increased input image resolution. Specifically, even with twice higher image resolution, the inference of ViTPose (\emph{ViT-S/16}) is still faster than that of EfficientDet (\emph{EfficientNet-B0}) that is 1/5 in size. Doubling the input resolution does lead to the increase of training latency by a factor of 4, but as stated in Requirement \#1 (\cref{sec:03_req:efficiency}), the increased training time is less likely to become as mission-critical as the inference time on a satellite with an onboard GPU. Moreover, ViTPose satisfies Requirement \#3 by design (\cref{sec:03_req:batch_agnostic}), completely lacking BN layers that could interfere with accurate online training on single images. 

Moving forward, the ViTPose models trained with the data augmentation proposed in \cref{sec:04_searching:augment} are referred to as SPNv3 for convenience. The default SPNv3 configuration is the ViT backbone pre-trained via DeiT III \citep{touvron_2022_eccv_deit3} on IN-22K then fine-tuned on IN-1K, patch size $P = 16$, and input resolution 448 $\times$ 448.

\subsubsection{Comparison on SPEED+}
\label{sec:05_exp_odd_latency:final_results:speedplus}

\begin{figure*}[!t]
    \centering
    \begin{subfigure}[b]{0.35\linewidth}
        \includegraphics[width=1.0\linewidth]{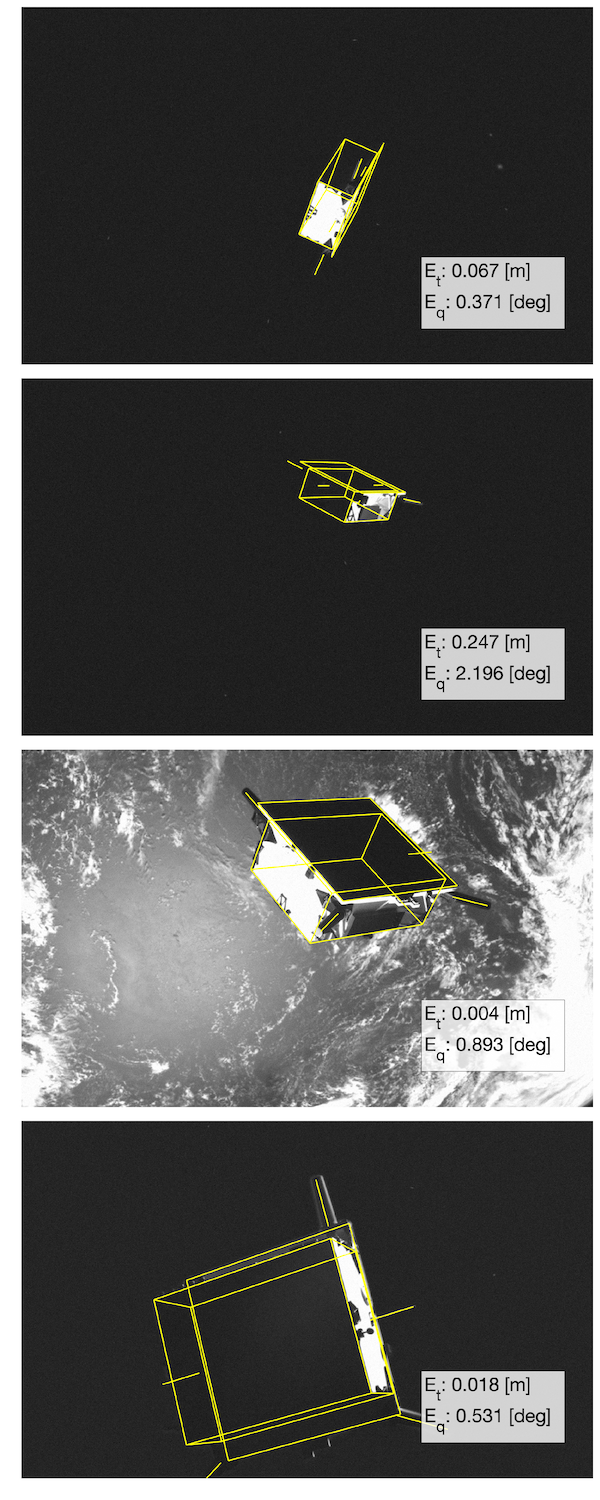}
        \caption{$\lightbox$}
    \end{subfigure}
    \begin{subfigure}[b]{0.35\linewidth}
        \includegraphics[width=1.0\linewidth]{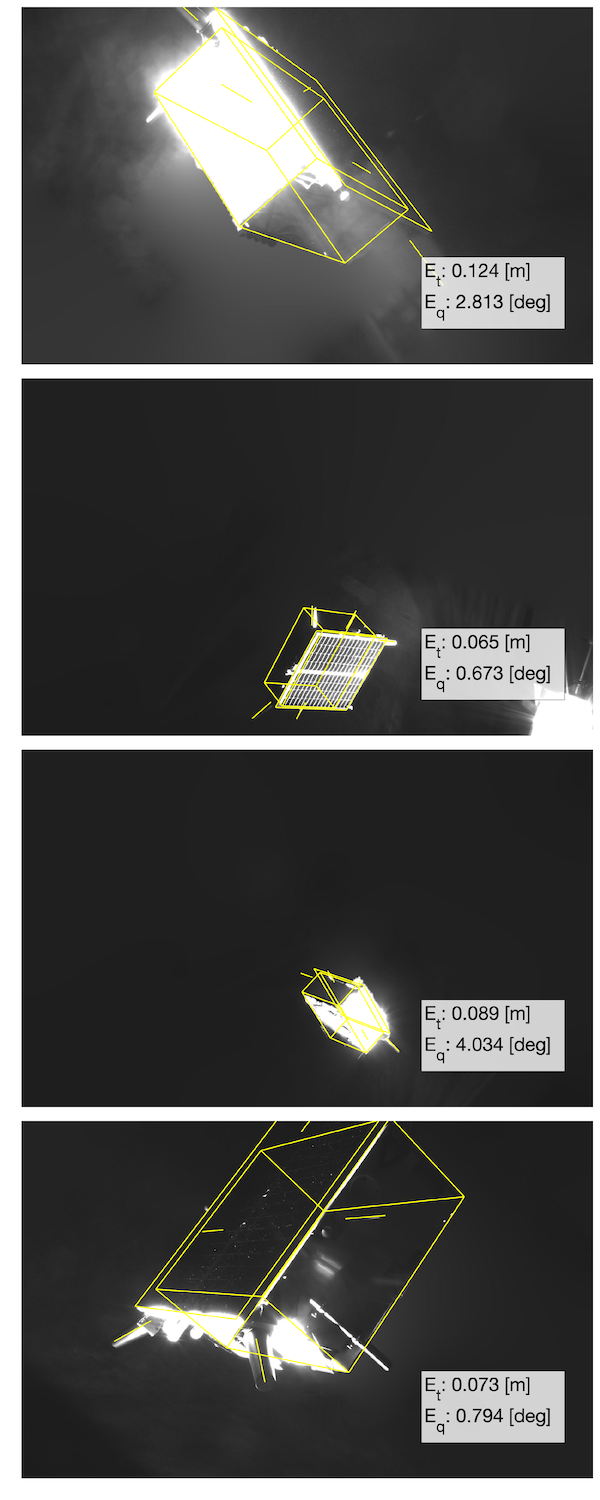}
        \caption{$\sunlamp$}
    \end{subfigure}
    \begin{subfigure}[b]{0.283\linewidth}
        \includegraphics[width=1.0\linewidth]{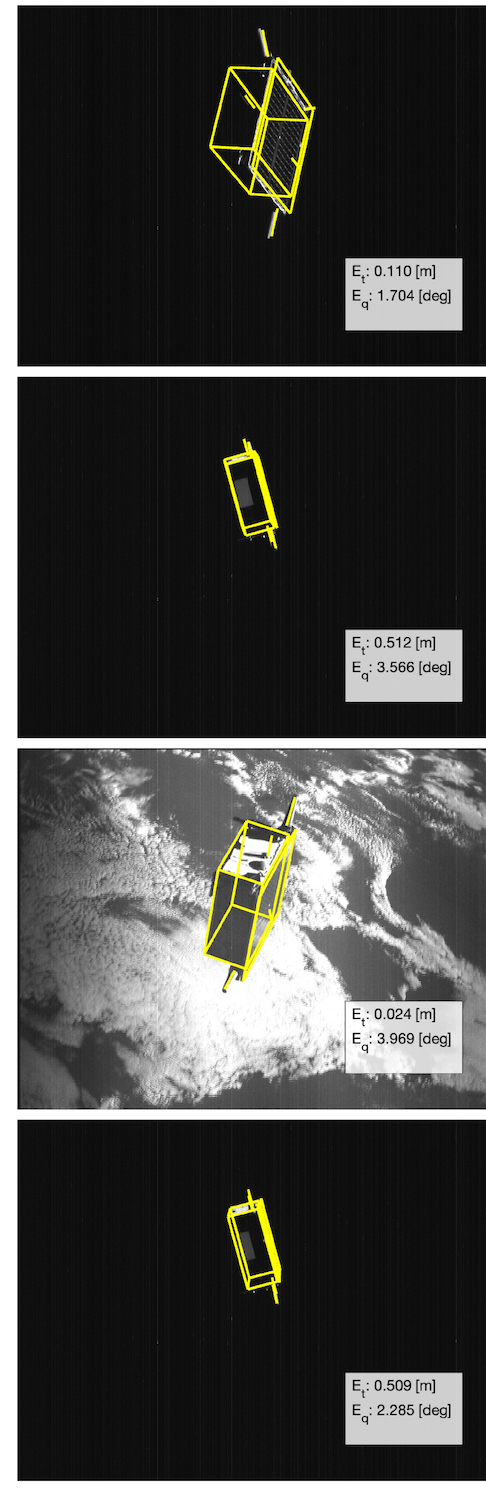}
        \caption{$\prisma$}
    \end{subfigure}
    \caption{Projection of the Tango wireframe model based on the poses predicted by the ensemble of SPNv3-S@448 on the sample SPEED+ HIL and $\prisma$ images.}
    \label{fig:05:final_results:visualizaiton}
\end{figure*}

\Cref{tab:05:final_results:speedplus} compares the final performances of SPNv3-S, a small variant that takes the ViT-S/16 backbone (22.7M parameters), against the top entries of SPEC2021 \citep{park_2023_acta_spec2021} and other state-of-the-art models introduced during the post-mortem competition following SPEC2021 and others. It can be seen that, with a higher input resolution at 448 $\times$ 448, SPNv3-S already outperforms the top entries of the $\lightbox$ category. Moreover, ensembling the output heatmap predictions from three models trained with different random seeds further improves the overall performance, achieving $E_\text{q} = 2.69^\circ$ on $\lightbox$ and $E_\text{q} = 3.89^\circ$ on $\sunlamp$. Some of the pose predictions by the ensemble are visualized in \cref{fig:05:final_results:visualizaiton} via reprojection of the Tango wireframe model. Note that the winning teams for respective HIL categories---TangoUnchained and lava1302---adopted adversarial training, directly utilizing the unlabeled images of the HIL domains. On the other hand, the proposed SPNv3 does \emph{not} utilize any information about the HIL domain during the training phase. The performance of SPNv3-S is also comparable to those of the post-mortem entries, surpassing FA-VAE \citep{liu_2024_cja_favae} with 41.6M parameters (approximated from its DarkNet-53 backbone \citep{redmon_2018_yolov3}) in both categories and \citet{chen_2024_sensors_homography}, which adopts semi-supervised domain adaptation, in the $\lightbox$ category. The performances of different SPNv3 variants are also comparable to that of EagerNet \citep{ulmer_2023_iros_eagernet}, which does not involve the HIL domain images during training. Note that EagerNet uses a pre-trained ConvNeXt-Base \citep{liu_2022_cvpr_convnext} network with 89M parameters and generates multiple pose hypotheses from dense predictions of model coordinates and predicted errors, which are further refined iteratively using a region-based approach. On the other hand, SPNv3-S is smaller and simpler, predicting keypoint locations from lower-resolution heatmaps after just a single forward propagation.

\begin{table*}[!t]
\caption{Performance of SPNv3 models and state-of-the-art. Mean{\footnotesize{(std.~dev.)}} denotes the average performance over 3 training sessions with different random seeds. The boldface denotes the best performance in each group.}
\label{tab:05:final_results:speedplus}
\centering
\footnotesize
\tabcolsep=0.15cm
\begin{tabular}{lccccccc}
\toprule
\multirow{2}{*}{Team/Model} & \multirow{2}{*}{\makecell{Num. \\ Params.}} & \multicolumn{3}{c}{$\lightbox$} & \multicolumn{3}{c}{$\sunlamp$} \\
\cmidrule(lr){3-5} \cmidrule(lr){6-8}
& & $\bar{E}_\textrm{t}^*$ [-] & $E_\textrm{q}^*$ [${}^\circ$] & $E_\textrm{pose}^*$ [-] & $\bar{E}_\textrm{t}^*$ [-] & $E_\textrm{q}^*$ [${}^\circ$] & $E_\textrm{pose}^*$ [-] \\
\midrule
\multicolumn{8}{l}{\underline{SPEC2021 Top-3 \citep{park_2023_acta_spec2021}}} \\
TangoUnchained \citep{park_2023_acta_spec2021}\textsuperscript{*} & - & \bfseries 0.018 & \bfseries 3.187 & \bfseries 0.073 & 0.015 & 4.299 & 0.090 \\ 
VPU \citep{perez_villar_2022_eccv_vpu}\textsuperscript{*} & - & 0.022 & 4.577 & 0.101 & 0.012 & 2.827 & 0.061 \\ 
lava1302 \citep{wang_2023_taes_lava1302}\textsuperscript{*} & - & 0.048 & 6.664 & 0.165 & \bfseries 0.011 & \bfseries 2.728 & \bfseries 0.059 \\ 
\midrule
\multicolumn{8}{l}{\underline{Post-mortem \& misc.}} \\
SPNv2 \citep{park_2023_asr_spnv2} & 52.5M & 0.025 & 5.577 & 0.122 & 0.026 & 9.788 & 0.197 \\
FA-VAE \citep{liu_2024_cja_favae} & 41.6M\textsuperscript{1} & 0.027 & 4.939 & 0.114 & 0.028 & 5.185 & 0.118 \\
\citet{chen_2024_sensors_homography}\textsuperscript{*} & - & 0.018 & 2.865 & 0.068 & 0.014 & 2.750 & 0.062 \\
EagerNet \citep{ulmer_2023_iros_eagernet} & 88.6M\textsuperscript{2} & \bfseries 0.009 & \bfseries 1.748 & \bfseries 0.039 & \bfseries 0.013 & \bfseries 2.664 & \bfseries 0.059 \\
\midrule
\multicolumn{8}{l}{\underline{Proposed}} \\
SPNv3-S & 22.7M & \mnstd{0.016}{0.001} & \mnstd{3.044}{0.234} & \mnstd{0.069}{0.004} & \mnstd{0.020}{0.000} & \mnstd{4.371}{0.314} & \mnstd{0.097}{0.006} \\ 
SPNv3-S\textsuperscript{\textdagger} & 22.7M & 0.017 & 2.689 & 0.064 & 0.020 & 3.883 & 0.088 \\ 
SPNv3-M & 39.6M & \mnstd{0.014}{0.000} & \mnstd{2.595}{0.123} & \mnstd{0.060}{0.002} & \mnstd{0.018}{0.001} & \mnstd{4.158}{0.254} & \mnstd{0.090}{0.005} \\ 
SPNv3-M\textsuperscript{\textdagger} & 39.6M & 0.014 & 2.396 & 0.056 & 0.017 & 3.765 & 0.082 \\ 
SPNv3-B & 86.3M & \mnstd{0.013}{0.001} & \mnstd{2.183}{0.129} & \mnstd{0.051}{0.003} & \mnstd{0.016}{0.001} & \mnstd{3.612}{0.164} & \mnstd{0.079}{0.003} \\ 
SPNv3-B\textsuperscript{\textdagger} & 86.3M & \bfseries 0.012 & \bfseries 2.033 & \bfseries 0.047 & \bfseries 0.015 & \bfseries 3.400 & \bfseries 0.074 \\ 
\bottomrule
\multicolumn{8}{l}{\footnotesize\textsuperscript{*} Included HIL domain images into training} \\
\multicolumn{8}{l}{\footnotesize\textsuperscript{\textdagger} From ensemble of heatmap predictions from the models of three training sessions} \\
\multicolumn{8}{l}{\footnotesize\textsuperscript{1} Approx.~from DarkNet-53 backbone \citep{redmon_2018_yolov3} ~~~ \textsuperscript{2} Approx.~from ConvNeXt-B backbone \citep{liu_2022_cvpr_convnext}}
\end{tabular}
\end{table*}

In the end, the ensemble performance with 448 $\times$ 448 inputs would rank SPNv3-S at first place in the $\lightbox$ category and third place in the $\sunlamp$ category of SPEC2021 \citep{park_2023_acta_spec2021}. Indeed, running an ensemble of multiple models may be computationally expensive especially onboard the satellite avionics. However, considering that the inference of SPNv3-S with 448 $\times$ 448 inputs only takes about 34 ms on a GPU of an NVIDIA Jetson Nano (see \cref{tab:05:ablation:in_resolution}), it may be possible to run an ensemble of a few models in real-time during RPOD. The largest variant of equivalent size, SPNv3-B with 86.3M parameters, marginally improves the accuracy, nearly matching that of EagerNet on $\lightbox$ yet still remaining in third place on $\sunlamp$. However, such a difference in performance between SPNv3-S and SPNv3-B would likely not yield significant improvement that justifies a near quadruple increase in size when processed by an onboard navigation filter.

Finally, while this article focuses on the OOD robustness and onboard computational capability of SPNv3 as an independent module, the integrated performance of SPNv3 and AUKF, both with and without online supervised training, is available for interested readers in \citet{park_2024_phd}.

\subsection{Limitations}
\label{sec:05_exp_odd_latency:limitations}

The proposed SPNv3 based on ViTPose is not without limitations. One major limitation is its dependence on a pre-trained ViT backbone. More specifically, \cref{tab:05:ablation:transfer} reveals that the OOD robustness of the downstream pose estimation task depends not just on what dataset the ViT backbone is pre-trained on but also on how the pre-training was conducted. This observation implies that those who cannot afford the computational power nor the expertise of ImageNet pre-training are restricted in exploring and fine-tuning different architectural options. In fact, the dependence on ImageNet pre-training prevents this work from exploring other aspects of ViT backbones, e.g., varying the patch sizes ($P$). Furthermore, the optimally pre-trained backbone may not exist, in which case the sub-optimally pre-trained alternative may have to be used for training on SPEED+. In response to such cases, the authors' previous work \citep{park_2024_icra_ost} shows how one can further train the sub-optimal NN onboard the satellite avionics in real-time during RPOD. However, if a NN can be trained according to the recipe analyzed in this section to be as OOD robust as possible, the online training on satellite avionics may be minimized or skipped altogether.
\section{Conclusions}
\label{sec:07_conclusion}

This paper presented SPNv3, a computationally efficient and robust NN model based on the ViT architecture. SPNv3 is trained exclusively on computer-generated synthetic images to achieve state-of-the-art robustness on unknown spaceborne imagery. Then, it is tested exhaustively on the hardware-in-the-loop images from a robotic facility that can simulate high-fidelity spaceborne illumination conditions. Extensive analyses are performed to identify data augmentation, transfer learning, NN size and input image resolution as key aspects of the design and training that contribute the most to its OOD robustness. In the end, a small SPNv3-S with 22.7M parameters can perform inference on larger 448 $\times$ 448 input images at nearly 30Hz on a restrictive GPU, which is well above the nominal update frequency of modern satellite guidance, navigation and control systems. Therefore, this paper successfully demonstrated that SPNv3 is flight ready, capable of achieving state-of-the-art robustness while simultaneously meeting the prescribed computational requirements.

The immediate next step is to address the requirement of power consumption by minimizing the memory footprint of the NN while maintaining the OOD robustness of larger SPNv3 variants. Candidate methods include NN weight pruning and quantization to reduced-precision formats. Furthermore, the OOD robustness evaluation in this work relies on the SPEED+ dataset, which is limited to a 10 m separation between the camera and the target due to the hardware restrictions. Therefore, the OOD robustness must be evaluated at a farther distance which predicates on improved calibration of the robotic testbed.

\section*{Funding Sources}

This work is supported by the US Space Force SpaceWERX Orbital Prime Small Business Technology Transfer (STTR) contract number FA8649-23-P-0560 awarded to TenOne Aerospace in collaboration with SLAB. The authors would like to thank OHB Sweden for the PRISMA images used in this work. The authors also thank Dr.~Justin Kruger at Stanford's Space Rendezvous Laboratory for providing an image of angles-only navigation from the NASA Starling mission.


\bibliography{bibliography}

\end{document}